%% file: acl_latex.tex
\documentclass[11pt]{article}

\usepackage[preprint]{acl}

\usepackage{times}
\usepackage{latexsym}

\usepackage[T1]{fontenc}

\usepackage[utf8]{inputenc}

\usepackage{microtype}

\usepackage{inconsolata}

\usepackage{graphicx}

\usepackage{placeins}

\usepackage{afterpage}
\usepackage{microtype}
\usepackage{graphicx}
\usepackage{subcaption}
\usepackage{booktabs} 
\usepackage{multicol}
\usepackage{multirow}
\usepackage{xcolor, colortbl}
\usepackage{listings}
\usepackage{placeins}
\usepackage{hyperref}
\usepackage{algorithmic}
\usepackage{algorithm}
\usepackage{amsmath}
\usepackage{amssymb}
\usepackage{mathtools}
\usepackage{amsthm}

\title{SFMP: Fine-Grained, Hardware-Friendly and Search-Free Mixed-Precision Quantization for Large Language Models}

\author{
\textbf{Xin Nie}\textsuperscript{1} \quad
 \textbf{Haicheng Zhang}\textsuperscript{1} \quad
 \textbf{Liang Dong}\textsuperscript{1} \quad
 \textbf{Beining Feng}\textsuperscript{1} \quad
 \textbf{Jinhong Weng}\textsuperscript{1} \quad \\
 \textbf{Guiling Sun}\textsuperscript{1} \quad \\
 \textsuperscript{1} College of Electronic Information and Optical Engineering, Nankai University\\
 \texttt{\{2120240458\}@mail.nankai.edu.cn}
}

\begin{document}
\maketitle
\begin{abstract}
  Mixed-precision quantization is a promising approach for compressing large language models under tight memory budgets. However, existing mixed-precision methods typically suffer from one of two limitations: they either rely on expensive discrete optimization to determine precision allocation, or introduce hardware inefficiencies due to irregular memory layouts. We propose \textbf{SFMP}, a search-free and hardware-friendly mixed-precision quantization framework for large language models. SFMP defines a theoretically simple objective for mixed-precision quantization and is built upon three novel ideas: 1) \textit{Block-wise mixed-precision}, enabling fine-grained precision within weight matrices while remaining hardware-friendly; 2) \textit{Row-column weight reordering}, which improves structural alignment via row and column reordering, incurring only a small activation reordering overhead during inference; 3) \textit{Unified GEMM kernel}, which supports block-wise mixed-precision GEMM at arbitrary average bit-widths. Extensive experiments demonstrate that SFMP outperforms state-of-the-art layer-wise mixed-precision methods under the same memory constraints, while significantly reducing quantization cost and improving inference efficiency. Code is available at \url{https://github.com/Nkniexin/SFMP}.
\end{abstract}

\input{_introduction}
\input{_related_works}

\input{_methods}

\input{_experiments}

\input{_analysis_ablation_study}

\input{_conclusion}

\newpage

\input{_limitation}

\section*{Ethical Considerations}
This paper presents work whose goal is to advance the field of Machine
Learning. There are many potential societal consequences of our work, none
which we feel must be specifically highlighted here.

\section*{Use of AI Assistance}
AI assistants were used solely for academic writing support, including language polishing, sentence refinement, and grammatical revision of the manuscript. They were not involved in the conception of research ideas, algorithm development, experimental design, or result analysis. All core contributions, technical content, and conclusions were independently developed by the authors.

\bibliography{custom}
\input{_appendix}

\end{document}

%% file: _introduction.tex
\vspace{-0.3cm}
\section{Introduction}
\vspace{-0.1cm}
Weight quantization is an efficient approach to compressing large language models (LLMs). It requires no modifications to the model architecture and directly maps high-precision continuous weights to a discrete space, reducing the average bits of model parameters, which enables deployment of LLMs in memory-constrained edge scenarios \cite{zhang2024edge,hosseinzadeh2025dilemma,husom2025sustainablellm}. Existing methods \cite{frantar2022gptq,xiao2023smoothquant,lin2024awq,kim2023squeezellm,liu2024spinquant} achieve near-lossless compression at 8-bit precision and only incur 1--3\% accuracy loss at 4-bit.


\begin{figure}[!t]
\centering
\includegraphics[width=\linewidth]{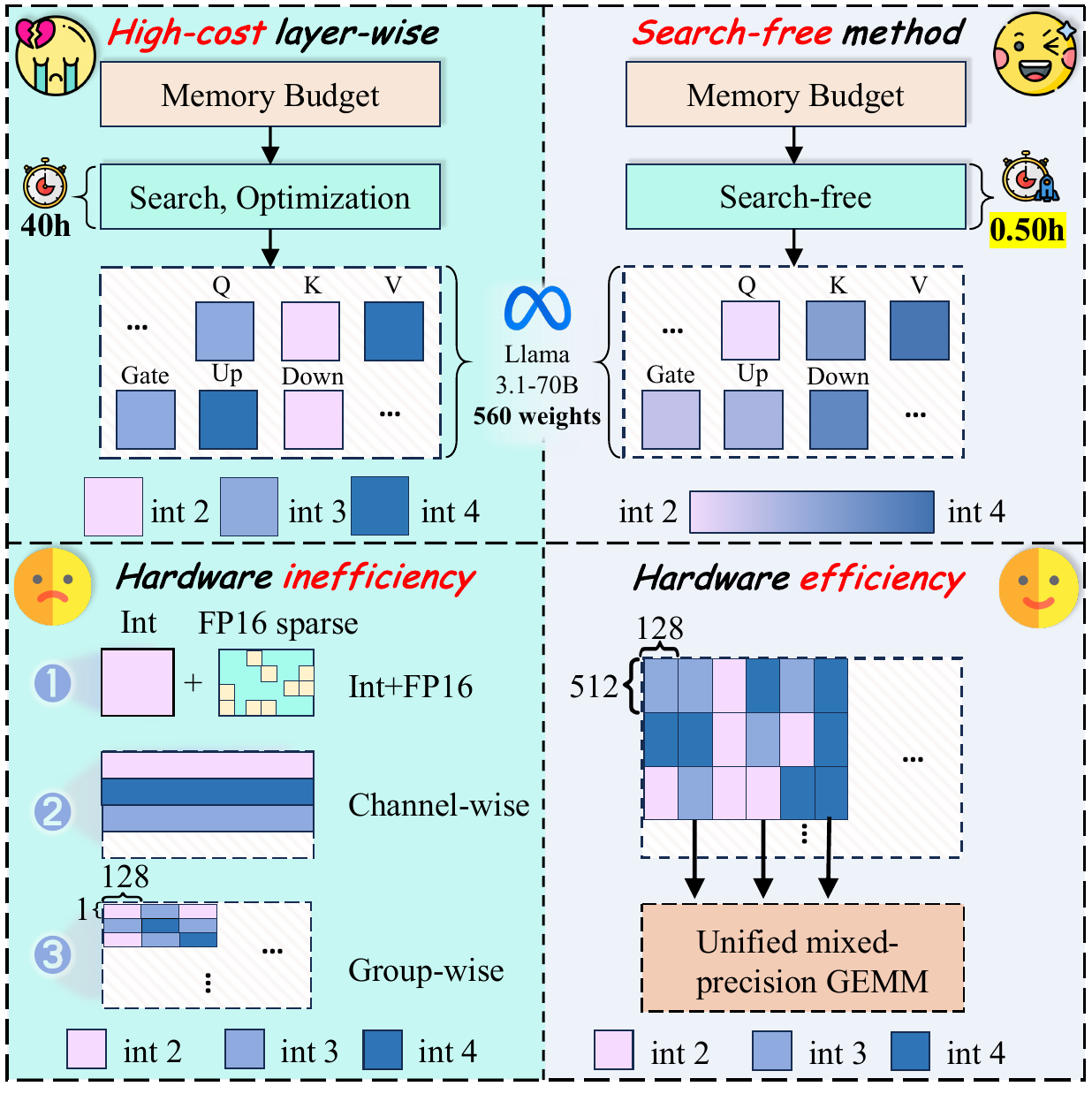} 
\caption{Comparison of SFMP with other mixed-precision quantization methods.}
\label{figure:motivation}
\vspace{-0.6cm}
\end{figure}

For ultra-large models, such as LLaMA3.1-70B, assigning a uniform integer bit-width to all linear layers limits the choices of model size and cannot adapt to diverse memory budgets. To address this limitation, layer-wise mixed-precision quantization \cite{lee2025amq,cheng2025signroundv2,liu2025flexquant,dong2020hawqv2} assigns different integer bit-widths to each linear layer, enabling flexible compression under given memory constraints. However, from an optimization perspective, layer-wise mixed-precision quantization constitutes a constrained integer programming problem (ILP), which is known to be NP-hard \cite{hochba1997approximationnphard}. For example, for LLaMA3.1-70B (560 weight matrices) with candidate bit-widths $\{2,3,4\}$, the search space is $3^{560}$. Existing methods typically transform the problem to fit off-the-shelf ILP solvers \cite{bellman1966dynamic,wolsey2020integer} or heuristic algorithms \cite{deb2002nsgaii,kirkpatrick1983anneal} to obtain a relatively good solution in a short time. Even so, as shown in Fig.~\ref{figure:motivation}, state-of-the-art layer-wise mixed-precision methods AMQ \cite{lee2025amq} still require 44 hours to search for the bit allocation of LLaMA3.1-70B. This raises a question: \textbf{``Under a given memory budget, can we design a strategy to obtain a near-optimal bit allocation without any search or solver-based optimization?"}

Beyond layer-wise mixed-precision quantization, many methods introduce finer-grained strategies, such as channel-wise \cite{chen2025channelwise,wang2024outliertune}, group-wise \cite{huang2024slim,hooper2025fgmp}, or even element-wise schemes \cite{kim2023squeezellm,zhao2025ganq,li2023llm-mq} in a weight matrix. Although such strategies can further improve model accuracy, they induce irregular memory access patterns and incur substantial overhead in weight packing and unpacking, which significantly degrades inference performance. Some approaches \cite{kim2023squeezellm,chen2025channelwise} attempt to mitigate this inference speed degradation with custom compute kernels \cite{li2024fast,qin2020sigma,liu2025paretoq}. However, the resulting speedup is limited. For example, our empirical study in Fig.~\ref{figure:empirical_study_slim} shows that the group-wise mixed-precision method SliM-LLM \cite{huang2024slim} suffers 30--50\% lower inference throughput than GPTQ \cite{frantar2022gptq} at the same average bit. Additionally, combining these fine-grained schemes with layer-wise mixed-precision further complicates the discrete optimization problem. This raises another important question: \textbf{``Can we design a quantization format that achieves fine-grained mixed-precision while remaining hardware-friendly?"}

In this paper, to address these limitations, we propose \textbf{SFMP}, a \textbf{Search-free Mixed-precision} framework. SFMP eliminates the need to solve complex integer programming problems under memory constraints, reducing the time required for compressing LLMs. For example, SFMP completes bit allocation for LLaMA3.1-70B in just 30 minutes. Moreover, SFMP is built upon three key ideas: 1) \textbf{Block-wise mixed-precision}: a format achieves fine-grained mixed-precision while remaining hardware-friendly; 2) \textbf{Row-column weight reordering}: weights are reordered along rows and columns to align with the block-wise format, incurring only a small activation reordering overhead during inference; 3) \textbf{Unified GEMM kernel}: for weight matrices composed of heterogeneous-precision blocks, we propose a unified kernel for memory layout and mixed-precision GEMM execution. Fig.~\ref{figure:motivation} shows the overview of SFMP.

%% file: _related_works.tex
\vspace{-0.2cm}
\section{Related Works}
\textbf{Salience-Aware Mixed-Precision Quantization.} Weight salience is widely used to guide mixed-precision quantization. GPTQ~\cite{frantar2022gptq} reorders the weight matrix column-wise according to the diagonal entries of the Hessian, prioritizing columns associated with larger diagonal values during quantization. SqueezeLLM \cite{kim2023squeezellm} computes the global Fisher Information Matrix \cite{ly2017fisherinformation} and retains a small set of salient weights in high precision. BiLLM \cite{huang2024billm} observes that salience often concentrates along specific rows or columns and reduces quantization error through column-wise partitioning. Similarly, Slim-LLM \cite{huang2024slim} exploits row-wise salience by introducing group-wise mixed-precision quantization, achieving improved accuracy over fixed-precision schemes under the same average bit-width.


\begin{figure}[!t]
\centering
\includegraphics[width=\linewidth]{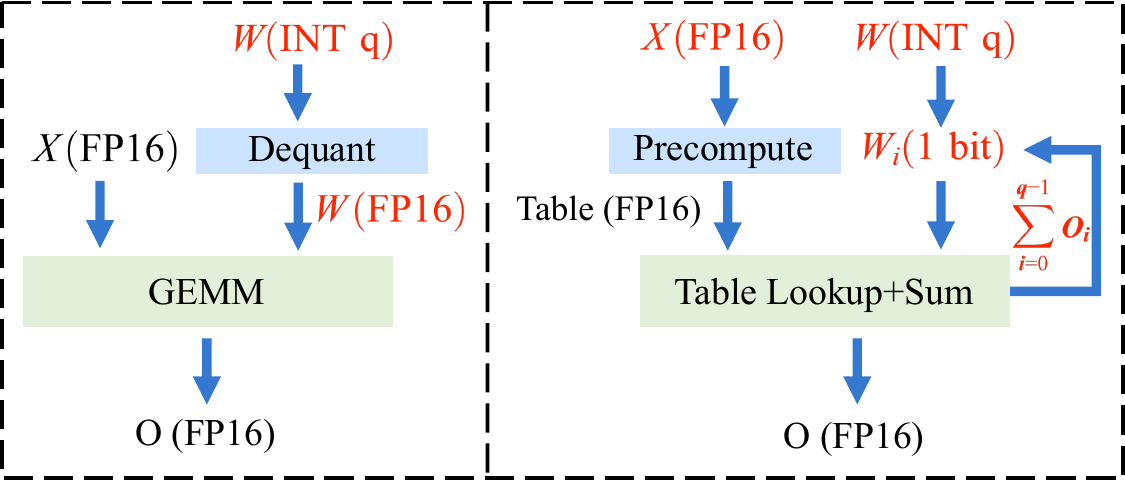} 
\caption{Comparison of two GEMM computation paradigms: Left) dequant-based GEMM; Right) one-bit LUT-based GEMM.}
\label{figure:one-bit-lut-gemm}
\vspace{-1cm}
\end{figure}

\textbf{One-bit LUT-Based GEMM.} Some prior works \cite{Weitmac, park2025figlut,park2022lut,you2024shiftaddllm,ganji2023deepgemm,park2025anybcq} introduce a dequantization-free compute paradigm for FP-INT GEMM. As shown in Fig.~\ref{figure:one-bit-lut-gemm}, a $q$-bit quantized weight matrix is decomposed into $q$ one-bit matrices. For an activation vector of group size $g$, the dot products between the activation and all $2^g$ possible binary patterns are precomputed and stored in a lookup table (LUT). Consequently, the GEMV operation between the activation vector and the one-bit weight matrix can be replaced by LUT accesses and accumulation. This paradigm eliminates the overhead of weight unpacking at runtime, particularly for ultra low-bit quantization. Moreover, it provides a unified computation kernel: The GEMM computation between activation and a weight matrix of arbitrary integer bit-width can be expressed as a linear combination of one-bit GEMMs. Owing to its LUT-dominated execution, this paradigm has been demonstrated to be energy-efficient \cite {jeon2020biqgemm,cicek2022energy,kim2025accelerating}, making it especially suitable for edge devices. A detailed description of the computation flow is provided in Appendix~\ref{section:details_about_one-bit_gemm}.

%% file: _methods.tex
\begin{figure*}[bt!]
    \centering
    \includegraphics[width=\textwidth]{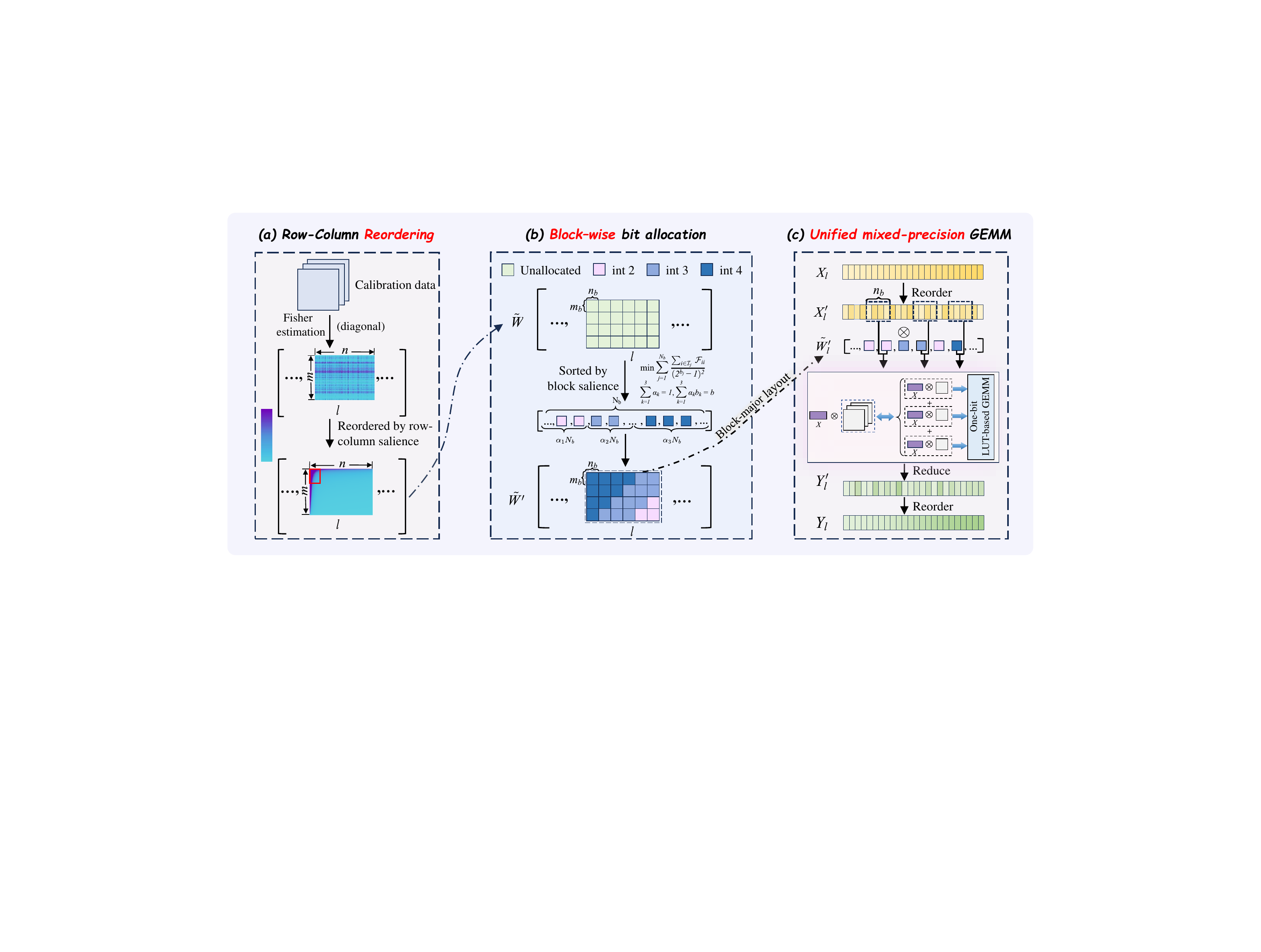}
    \caption{Pipeline of SFMP. a) Row-column reordering to aggregate salient weights; b) Block-wise mixed-precision allocation; c) Activation reordering followed by unified mixed-precision GEMM for matrix multiplication.}
    \label{figure:pipeline}
    \vspace{-0.5cm}
\end{figure*}

\section{SFMP}
\vspace{-0.1cm}
\subsection{Preliminaries}
As pointed out in prior works~\cite{kim2023squeezellm,zhao2025ganq}, the impact of quantized weights on the model output can be estimated via Taylor expansion. Assuming the model has converged, the loss variation $\Delta \mathcal{L}$ induced by a perturbation from $W$ to $W'$ can be approximated by a second-order expansion:
\vspace{-0.1cm}
\begin{equation}
\Delta \mathcal{L} \approx (W - W')^\top H (W - W'),
\end{equation}
\vspace{-0.1cm}
where $H$ denotes the Hessian of the global loss with respect to weights. Since computing the full Hessian is intractable for large-scale models, we approximate it using the Fisher Information Matrix \cite{ly2017fisherinformation}, $H \approx \mathcal{F} = \mathbb{E}[g g^\top]$, where $g$ denotes the gradient of the loss with respect to the weights, and the expectation is taken over a small calibration dataset. Following prior work \cite{kim2023squeezellm}, we adopt a diagonal approximation by ignoring cross-weight interactions. Therefore, the quantization-induced perturbation of the loss can be written as:
\begin{equation}
\label{equation:model_loss}
\Delta\mathcal{L} \approx \sum_{i=1}^{N}\mathcal{F}_{ii}(W_i - W_i')^2,
\end{equation}
where $N$ denotes the total number of scalar weights in the model. Detailed derivations are provided in Appendix~\ref{section:derivation_of_global_salience}.

\vspace{-0.1cm}
\subsection{Objective of SFMP}
\label{subsection:objective}

From Eq.~\ref{equation:model_loss}, we formulate the mixed-precision bit allocation problem as follows. Given a candidate set of integer bit-widths $\mathcal{B} = \{ b_1, b_2, \dots, b_q \}$ with $b_i \in \mathbb{Z}_{>0}$ and $b_1 < b_2 < \cdots < b_q$, we treat each weight element as the minimal quantization unit. For a target average bit-width $b \in \mathbb{R}_{>0}$, the goal is to find coefficients $\{\alpha_1, \alpha_2, \dots, \alpha_q\}$ such that
\vspace{-0.1cm}
\begin{equation}
\begin{aligned}
\min \quad & \Delta \mathcal{L} \approx  \sum_{i=1}^{N} \mathcal{F}_{ii} (W_i - W_i')^2, \\ 
\text{s.t.} \quad & \sum_{k=1}^q \alpha_k = 1, \sum_{k=1}^q \alpha_k b_k = b.
\end{aligned}
\end{equation}

To make the objective tractable, we consider the expected loss increase under stochastic quantization. Therefore, the objective becomes 
\begin{equation}
\mathbb{E}[\Delta \mathcal{L}] \approx \sum_{i=1}^{N} \mathcal{F}_{ii} \mathbb{E}[(W_i - W_i')^2].
\end{equation}
Specifically, we assume that each weight is quantized independently under a uniform quantizer with quantization step $\delta$, and the dynamic range $(W_{\max} - W_{\min})$ is approximately constant across weights. Under this assumption, the quantization error can be modeled as a uniform random variable $x \sim \mathcal{U}(-\delta/2, \delta/2)$, leading to
\begin{equation}
\mathbb{E}[(W_i - W_i')^2] = \frac{1}{\delta}\int_{-\delta/2}^{\delta/2}x^2dx = \frac{\delta^2}{12},
\end{equation}
where 
$\delta = \frac{W_{\max} - W_{\min}}{2^{b_i} - 1}.$

Substituting the above relation into the Fisher-weighted objective, where the expected quantization error is determined by the bit-width $b_i$, yields the following bit allocation objective:
\begin{equation}
\label{equation:objective}
\begin{aligned}
\min \quad & \sum_{i=1}^{N} \frac{\mathcal{F}_{ii}}{(2^{b_i} - 1)^2} \\
\text{s.t.} \quad & \sum_{k=1}^q \alpha_k = 1, \sum_{k=1}^q \alpha_k b_k = b.
\end{aligned}
\end{equation}

Here, we present \textit{sorting-based} solutions for different sizes of the candidate bit-width set $\mathcal{B}$, avoiding iterative optimization.

\paragraph{Case 1: $|\mathcal{B}| = 2$.}
When $\mathcal{B} = \{b_1, b_2\}$, the coefficients $\alpha_1$ and $\alpha_2$ are uniquely determined by the constraints. The problem reduces to a special case of the 0-1 knapsack problem. Since assigning higher bit-widths to weights with larger $\mathcal{F}_{ii}$ always yields lower loss, the optimal solution is obtained by sorting $\mathcal{F}_{ii}$ in ascending order, assigning the smallest $\alpha_1$ portion to $b_1$, and the remaining $\alpha_2$ to $b_2$.

\paragraph{Case 2: $|\mathcal{B}| = 3$.}

When $\mathcal{B} = \{b_1, b_2, b_3\}$, we adopt a simple one-dimensional grid search over $\alpha_1$, as summarized in Alg.~\ref{alg:elementwise_b3}. Once $\alpha_1$ is fixed, $\alpha_2$ and $\alpha_3$ are uniquely determined, reducing the problem to the \textbf{Case 1}. In practice, a grid step size $\Delta$ of $0.01$ achieves a good trade-off between efficiency and performance. A further analysis of $\Delta$ is provided in Appendix~\ref{section:analysis_of_search_step}.

\paragraph{Case 3: $|\mathcal{B}| > 3$.}
In principle, the \textbf{Case 2} can be extended to multi-dimensional grid search. However, the search complexity grows exponentially with $|\mathcal{B}|$, while the performance gain is marginal. As shown in Appendix~\ref{section:analysis_of_candidate_bitwidth}, increasing the number of candidate bit-widths beyond three brings negligible improvement compared to the $|\mathcal{B}|=3$ setting.

\begin{algorithm}[tb]
  \caption{Element-wise Mixed-Precision Allocation ($|\mathcal{B}|=3$)}
  \label{alg:elementwise_b3}
  \begin{algorithmic}
    \STATE {\bfseries Input:} 
        $\{\mathcal{F}_{ii}\}_{i=1}^N$, target bit-width $b$, candidate bits $\{b_1,b_2,b_3\}$, step size $\Delta$
    \STATE {\bfseries Output:} bit assignment $\{b_i\}_{i=1}^N$

    \STATE Sort indices of $\{\mathcal{F}_{ii}\}$ in ascending order

    \FOR{$\alpha_1 = 0$ to $1$ with step $\Delta$}
        \STATE $\alpha_2 = \dfrac{b - b_3 - (b_1 - b_3)\alpha_1}{b_2 - b_3}$

        \STATE $\alpha_3 = 1 - \alpha_1 - \alpha_2$

        \STATE $n_k = \lfloor \alpha_k N \rfloor,\quad k \in \{1,2,3\}$

        \STATE $b_i =
        \begin{cases}
        b_1, & i \le n_1 \\
        b_2, & n_1 < i \le n_1 + n_2 \\
        b_3, & \text{otherwise}
        \end{cases}$
        
        \STATE $\mathcal{L} = \sum_{i=1}^{N} \frac{\mathcal{F}_{ii}}{(2^{b_i}-1)^2}$
    \ENDFOR

    \STATE Return assignment with minimum $\mathcal{L}$
  \end{algorithmic}
\end{algorithm}

\vspace{-0.3cm}
\subsection{Block-Wise Mixed-Precision}
Although the element-wise bit allocation strategy in Alg.~\ref{alg:elementwise_b3} is theoretically accurate and simple to implement, it is inefficient for hardware execution. Weights assigned the same bit-width are scattered in a highly irregular spatial pattern within the weight matrix, hindering structured quantization schemes such as group quantization.

To balance quantization performance and hardware efficiency, we adopt a coarser yet structured block-wise bit allocation strategy, where bit-widths are assigned at the block level rather than to individual weights. Specifically, as shown in Fig.~\ref{figure:pipeline}(b), each weight matrix is partitioned into non-overlapping two-dimensional blocks of size $m_b \times n_b$. Leveraging the additive structure of the objective in Eq.~\ref{equation:objective}, we reformulate it as:

\begin{equation}
\min \sum_{j=1}^{N_b} \frac{\sum_{i \in \mathcal{I}_j} \mathcal{F}_{ii}}{(2^{b_j} - 1)^2} 
\end{equation}

where $\mathcal{I}_j$ denotes the index set of weights in the $j$-th block, and $N_b$ is the total number of blocks. Treating each block as the minimal unit for precision allocation, the strategy proposed in Section~\ref{subsection:objective} can be directly applied. The resulting block-wise scheme exhibits regular and hardware-friendly bit patterns. In practice, the block dimensions $(m_b, n_b)$ are chosen to balance fine-grained granularity and hardware characteristics. For example, on GPUs, we adopt block sizes such as $(256,128)$ or $(512,128)$ to match common GEMM tiling strategies and warp-level parallelism in CUDA.

\vspace{-0.1cm}
\subsection{Row-Column Weight Reordering}
As shown in Fig.~\ref{figure:pipeline}(a), it can be observed that the spatial distribution of the diagonal values of the Fisher information matrix exhibits strong structure along rows or columns. This distributional property has been exploited by prior works~\cite{huang2024billm,huang2024slim} to guide quantization, for example, by reordering weights column-wise. In our work, our block-wise bit allocation is spatially misaligned with this distribution. To address this mismatch, we further propose a bidirectional reordering strategy that reorganizes the weight matrix based on both row-wise and column-wise aggregated Fisher diagonal values.

Given a weight matrix $W_l \in \mathbb{R}^{m \times n}$, we define row- and column-wise salience by aggregating the diagonal entries of the Fisher information matrix:
\vspace{-0.1cm}
\begin{equation}
s_{l,\text{row}} = \sum_{i \in \mathcal{R}_l} \mathcal{F}_{ii}, \qquad
s_{l,\text{col}} = \sum_{i \in \mathcal{C}_l} \mathcal{F}_{ii},
\end{equation}
where $\mathcal{R}_l$ and $\mathcal{C}_l$ denote the index sets of weights in each row and column of $W_l$, respectively.

Row and column permutations are obtained by sorting these aggregated Fisher values in descending order:
$p_{l,\text{row}} = \operatorname{argsort}(s_{l,\text{row}}), 
p_{l,\text{col}} = \operatorname{argsort}(s_{l,\text{col}}).$
The corresponding permutation matrices are denoted by $P_{l,\text{row}}$ and $P_{l,\text{col}}$. The reordered weight matrix is given by:
\begin{equation}
\tilde{W}_l = P_{l,\text{row}} W_l P_{l,\text{col}}.
\end{equation}

As shown in Fig.~\ref{figure:pipeline}(a), the bidirectional reordering strategy spatially aggregate high-salience weights, improving the alignment between weight salience and block-wise bit allocation. Notably, the reordering is performed offline and incurs no runtime overhead. During inference, as shown in Fig.~\ref{figure:pipeline}(c), the same permutation can be equivalently applied to the activation, with negligible cost compared to GEMM computation.


\vspace{-0.2cm}
\subsection{Unified Mixed-Precision GEMM Kernel}
For our proposed block-wise mixed-precision format, adopting dequant-based operators introduces two major challenges: 1) conventional row-major or column-major storage layouts complicate weight packing and unpacking, as the block structure is not explicitly represented in memory. 2) mixed-precision formats require additional control-flow branching and multiple kernel variants in compute kernels (e.g., CUDA kernels), increasing implementation complexity.

As shown in Fig.~\ref{figure:pipeline}(c), to address challenge 1), we propose a block-major representation, where the quantized weight matrix is partitioned into blocks and organized in a block-major layout, enabling contiguous memory access within each block. To address challenge 2), we employ a unified GEMM kernel that processes all blocks regardless of their bit-widths. Specifically, each block is decomposed into one-bit components and computed via one-bit LUT-based GEMM, thereby eliminating explicit weight dequantization and precision-specific execution paths (see Appendix~\ref{section:cuda_implementataion} for detailed CUDA implementation).

A potential concern is that heterogeneous bit-widths may introduce load imbalance, with high-precision blocks dominating the overall latency. However, this is avoided in our design. Our kernel typically tiles the computation into thread blocks of size $[M_{\text{tile}}, K_{\text{tile}}]$, e.g., $[256,128]$. For a $4096 \times 4096$ matrix (e.g., the $q_{\text{proj}}$ in LLaMA3.1-8B), this results in $512$ thread blocks. From a hardware perspective, modern GPUs execute thread blocks via dynamic scheduling over Streaming Multiprocessors (SMs). Each thread block runs to completion on a single SM, while idle SMs continuously fetch new blocks from a global work queue. In our setting, the number of thread blocks is much larger than the number of SMs (e.g., 82 on RTX~3090 and 108 on A100), so each SM processes multiple blocks sequentially. In a multi-SM parallel setting, higher-precision blocks mainly affect the tail of overall execution, which constitutes only a small fraction of the total runtime. The overall latency is determined by the \textbf{aggregate workload}, rather than the highest-precision blocks.

%% file: _experiments.tex
\vspace{-0.2cm}
\section{Experiments}
We evaluate our method on several state-of-the-art pretrained models, including LLaMA3.1 8B and 70B \cite{dubey2024llama3}, Qwen3 8B, 14B and 32B \cite{yang2025qwen3}. The diagonal values of the Fisher Information Matrix are estimated using 1k samples from C4~\cite{raffel2020c4}. The block shape $(m_b,n_b)$ is (512, 128), where group quantization is applied along the $n_b$ dimension within each block. The candidate bit-width set is $\{2, 3, 4\}$. To ensure fair comparison, following AMQ \cite{lee2025amq}, we avoid introducing complex tricks and just adopt AWQ \cite{lin2024awq} as our quantization method. We compare our method against fixed-precision methods such as GPTQ and AWQ, element-wise mixed-precision method SqueezeLLM~\cite{kim2023squeezellm}, group-wise mixed-precision method SliM-LLM \cite{huang2024slim}, any-size methods such as BitStack \cite{wang2024bitstack} and AMQ. Our method is orthogonal to most quantization tricks, in Appendix~\ref{section:SFMP_QAT}, we further combine our approach with Quantization-Aware Training (QAT) method \cite{chen2024efficientqat}. All experiments are conducted on A100-80GB GPU.

We evaluate our method from multiple perspectives. For language modeling, we report perplexity on C4 and WikiText2 \cite{wikitext2}. For zero-shot evaluation, we use the LM Evaluation Harness \cite{gao2021framework} to evaluate six tasks, including ARC-Challenge, ARC-Easy \cite{clark2018think}, PIQA \cite{bisk2020piqa}, HellaSwag \cite{zellers2019hellaswag}, BoolQ \cite{clark2019boolq}, and WinoGrande \cite{sakaguchi2021winogrande}. We further evaluate 5-shot performance on MMLU \cite{hendrycks2020mmlu} and GSM8K \cite{cobbe2021gsm8k}. For inference performance, considering edge deployment scenarios, we report both kernel-level latency and end-to-end inference speed (tokens / s) when generating 128 tokens with batch size 1.

\vspace{-0.3cm}
\subsection{Main Results}
\textbf{SFMP vs. Any-Size Methods.} Table~\ref{tab:main_table} reports the model perplexity and zero-shot task accuracy under different memory budgets for models quantized with SFMP, AMQ, and BitStack. We report results under different \textbf{BPW} (bits per weight) settings. Across multiple model scales, SFMP consistently outperforms AMQ. The advantage of SFMP becomes more pronounced at extremely low precision (e.g., BPW=2.5), where it achieves the best average zero-shot accuracy among all methods. At BPW=3.5, SFMP retains 98.90\% of the average zero-shot performance of the BF16 LLaMA3.1-70B model. Moreover, Table~\ref{tab:mmlu_gsm8k} shows that on the 5-shot MMLU and GSM8K benchmarks, SFMP consistently outperforms AMQ across all model sizes. These results demonstrate that SFMP remains robust on challenging tasks and show that our strategy is more stable and effective than layer-wise mixed-precision methods, even without any search or optimization.

\renewcommand{\arraystretch}{0.75}
\begin{table}[thb!]

\vspace{-0.1cm}
\setlength{\tabcolsep}{1mm}
\resizebox{\columnwidth}{!}
{
\begin{tabular}{c|c|c|c||cc||c}
\midrule
\textbf{\large Model} &  \textbf{\begin{tabular}[c]{@{}c@{}}\textbf{\large{Mem.}}\\ \textbf{\large{(MB)}}\end{tabular}} & \textbf{\large BPW} &  \textbf{\large Method} &  \textbf{\large Wiki($\downarrow$)} &  \textbf{\large C4($\downarrow$)} & \textbf{\large Avg.($\uparrow$)} \\ \midrule
 & 15,317 & 16 & BF16 & 6.15 & 8.89 & 75.01 \\ \cmidrule{2-7}
 & \multirow{3}{*}{4,085}  & \multirow{3}{*}{2.5} 
 & BitStack & 23.28 & 38.23 & 58.19 \\
 & & & AMQ & 17.85 & 24.01 & 58.65\\
   &   &  & \textbf{SFMP} & \textbf{13.68} & \textbf{17.77} & \textbf{65.97}\\ \cmidrule{2-7}
 & \multirow{3}{*}{4,501} & \multirow{3}{*}{3.0} 
 & BitStack & 12.55 & 20.47 & 64.40\\
 & & & AMQ & 9.38 & 13.05 & 68.78 \\
   &   &   & \textbf{SFMP} & \textbf{8.65} & \textbf{12.04} & \textbf{69.92} \\ \cmidrule{2-7}
 & \multirow{3}{*}{4,917} & \multirow{3}{*}{3.5}
 & BitStack& 9.47& 15.29 & 68.59 \\ 
 & & & AMQ & 7.39 & 10.54 & 72.56 \\
   &   &   & \textbf{SFMP} & \textbf{7.30} & \textbf{10.38} &  \textbf{73.43} \\ \cmidrule{2-7}
 & \multirow{3}{*}{5,333} & \multirow{3}{*}{4.0} 
 & BitStack & 8.39 & 13.47 &  70.95 \\ 
 & & & AMQ & 6.86 & 9.79 & 73.46  \\
  \multirow{-16}{*}{\textbf{\large 8B}} &   &   & \textbf{SFMP} & \textbf{6.84} & \textbf{9.74} & \textbf{74.15}  \\ \midrule
 & 134,571 & 16 & BF16 & 2.81 & 7.11 & 80.96 \\ \cmidrule{2-7}
 & \multirow{3}{*}{24,411}& \multirow{3}{*}{2.5} 
 & BitStack & 7.55 & 12.92 & 74.51 \\
 & & & AMQ & 7.62 & 12.14 & 74.33 \\
   &   &   & \textbf{SFMP} & \textbf{7.24} & \textbf{10.07} & \textbf{74.60} \\ \cmidrule{2-7}
 & \multirow{3}{*}{28,491} & \multirow{3}{*}{3.0} 
 & BitStack & 6.38 & 11.21 & 76.30 \\
 & & & AMQ & 5.84 & 9.47 & 77.80\\
   &   &   & \textbf{SFMP} & \textbf{5.31} & \textbf{8.36} & \textbf{78.07} \\ \cmidrule{2-7}
 & \multirow{3}{*}{32,571} & \multirow{3}{*}{3.5} 
 & BitStack & 5.44 & 9.52 & 78.24 \\
 & & & AMQ & 4.26 & 8.20 &  79.11 \\
   &   &   & \textbf{SFMP} & \textbf{4.00} & \textbf{7.33} & \textbf{80.07} \\ \cmidrule{2-7}
 & \multirow{3}{*}{36,651} & \multirow{3}{*}{4.0} 
 & BitStack  & 4.98 & 8.92 & 79.17 \\
 & & & AMQ & 3.49 & 7.61 &  80.14\\
  \multirow{-16}{*}{\textbf{\large 70B}} &   &   & \textbf{SFMP} & \textbf{3.37} & \textbf{7.01} & \textbf{80.47}  \\ \bottomrule \bottomrule
\end{tabular}
}

\caption{Evaluation of Llama 3.1 8B/70B models compressed by SFMP, BitStack and AMQ at the BPW of 2.5, 3.0, 3.5 and 4.0, using a groupsize of 128, showing WikiText-2 and C4 dataset perplexity (PPL) alongside zero-shot tasks average accuracy. \textbf{BPW} denotes \textit{``bits per weight"}. Quantization scales and zero-points are stored in BF16. Detailed zero-shot accurcy is provided in Table~\ref{tab:llama3.1_full_results_mixed_precison}. }
\label{tab:main_table}

\vspace{-0.5cm}
\end{table}

\textbf{SFMP vs. Fixed-Precision Methods.} We compare SFMP with GPTQ, AWQ and SliM-LLM, which apply a uniform bit-width across all weight matrices. Table~\ref{tab:table_3} reports perplexity and average zero-shot accuracy on the BPW ranging from 2.25 to 4. Across all settings, SFMP consistently outperforms fixed-precision methods. Moreover, SFMP consistently outperforms the group-wise mixed-precision method SliM-LLM and provides greater flexibility in precision allocation, as SliM-LLM enforces fixed average bit-width across all weight matrices.

\begin{figure*}[!t]
\vspace{-0.2cm}
\centering
\includegraphics[width=\linewidth]{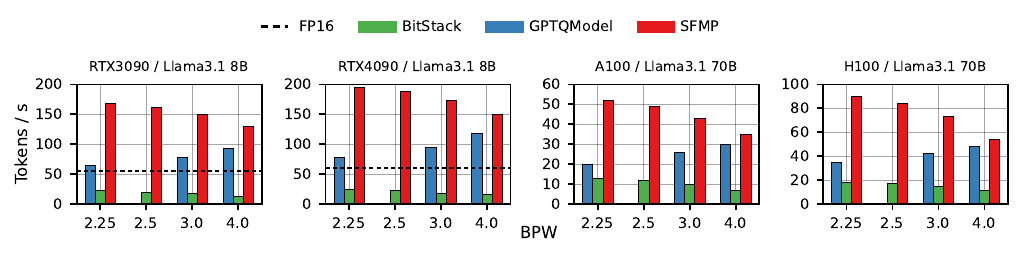} 
\vspace{-1cm}
\caption{End-to-end throughput (tokens/s) of generating a sequence length of 128 with batchsize of 1. BF16 inference of LLaMA3.1 70B is not feasible on single A100 and H100 due to memory constraints. GPTQModel uses Marlin~\cite{frantar2025marlin} for BPW=4, and Triton~\cite{tillet2019triton} for BPW=2 and 3. At BPW=4, we provide comparisons with TensorRT-LLM~\cite{tensorrt_llm} and ExLlama~\cite{exllama} in Table~\ref{tab:4bit_baselines}.}
\label{figure:inference_speed}
\vspace{-0.3cm}
\end{figure*}

\textbf{Inference Performance.} We evaluate the inference performance of SFMP across a wide range of hardware platforms. As shown in Fig.~\ref{figure:inference_speed}, fixed-precision uniform quantization framework GPTQModel \footnote{\url{https://github.com/ModelCloud/GPTQModel}} becomes slower as the BPW decreases. This counterintuitive behavior is caused by the increasing weight unpacking and dequantization overhead at low bit-width. In contrast, SFMP exhibits an increase in inference speed as the BPW reduces. This advantage stems from its one-bit LUT-based GEMM formulation, where the computational latency of the kernel scales approximately linearly with the BPW (see Fig.~\ref{figure:kernel_evaluation}). Moreover, the decomposition-based compression method BitStack suffers from repeated weight reconstruction during inference, leading to substantially worse performance, even compared to BF16.

\renewcommand{\arraystretch}{0.75}
\begin{table}[thb!]
\vspace{-0.2cm}

\setlength{\tabcolsep}{1mm}
\resizebox{\columnwidth}{!}
{
\begin{tabular}{c|c|c|c||cc||c}
\midrule
\textbf{\large Model} &  \textbf{\begin{tabular}[c]{@{}c@{}}\textbf{\large{Mem.}}\\ \textbf{\large{(MB)}}\end{tabular}} & \textbf{\large BPW} &  \textbf{\large Method} &  \textbf{\large Wiki($\downarrow$)} &  \textbf{\large C4($\downarrow$)} & \textbf{\large Avg.($\uparrow$)} \\ \midrule
 & 15,317 & 16 & BF16 & 6.15 & 8.89 & 75.01 \\ \cmidrule{2-7}
 & \multirow{3}{*}{3,877}  & \multirow{3}{*}{2.25} 
 & GPTQ$_{w2g128}$ & 232 & 165 & 38.55 \\
 & & & AWQ$_{w2g128}$ & 1.57e6 & 1.86e6 & 35.80 \\
 & & & SliM-LLM$_{g128}$ & 193 & 142 & 40.67 \\
 &  3,961 &  2.35 & \textbf{SFMP}$_{g128}$ & \textbf{24.57} & \textbf{28.92} & \textbf{60.80} \\ \cmidrule{2-7}
 & \multirow{4}{*}{4,501} & \multirow{4}{*}{3.0} 
 & GPTQ$_{w3}$ & 22.13 & 25.05 & 55.83 \\
 & & & AWQ$_{w3}$ & 16.06 & 19.79 & 64.61 \\
  & & & SqueezeLLM$_{w3}$ & 13.43 & 15.64 & 65.78 \\
   &   &   & \textbf{SFMP}$_{g128}$ & \textbf{8.65} & \textbf{12.04} & \textbf{69.92} \\ \cmidrule{2-7}
 & \multirow{5}{*}{4,709} & \multirow{5}{*}{3.25}
 & GPTQ$_{w3g128}$& 8.28 & 11.49 & 69.22 \\ 
 & & & AWQ$_{w3g128}$ & 8.23 & 11.58 & 70.72 \\
  & & & SliM-LLM$_{g128}$ & 8.17 & 11.25 & 70.31 \\
    & & & SqueezeLLM$_{0.45\%}$ & 7.95 & 11.39 & 70.97 \\
   &   &   & \textbf{SFMP}$_{g128}$ & \textbf{7.78} & \textbf{10.97} & \textbf{72.47} \\ \cmidrule{2-7}
 & \multirow{3}{*}{5,333} & \multirow{3}{*}{4.0} 
 & GPTQ$_{w4}$ & 7.5 & 10.38 & 71.46 \\ 
 & & & AWQ$_{w4}$ & 7.23 & 10.26 & 73.60 \\
  & & & SqueezeLLM$_{w4}$ & 7.17 & 10.11 & 73.17 \\
  \multirow{-16}{*}{\textbf{\large 8B}} &   &   & \textbf{SFMP}$_{g128}$ & \textbf{6.84} & \textbf{9.74} & \textbf{74.15} \\ \midrule
 & 134,571 & 16 & BF16 & 2.81 & 7.11 & 80.96 \\ \cmidrule{2-7}
 & \multirow{3}{*}{22,371}& \multirow{3}{*}{2.25} 
 & GPTQ$_{w2g128}$ & 113.22 & 131.9 & 40.02 \\
 & & & AWQ$_{w2g128}$ & 1.8e6 & 1.5e6 & 40.65 \\
 & & & SliM-LLM$_{g128}$ & 68.84 & 88.36 & 46.51 \\
   & 23,187  & 2.35 & \textbf{SFMP}$_{g128}$ & \textbf{8.17} & \textbf{11.42} & \textbf{72.65} \\ \cmidrule{2-7}
 & \multirow{4}{*}{28,491} & \multirow{4}{*}{3.0} 
 & GPTQ$_{w3}$ & 11.27 & 12.19 & 66.27 \\
 & & & AWQ$_{w3}$ & 10.86 & 11.74 & 68.84 \\
  & & & SqueezeLLM$_{w3}$ & 10.17 & 10.62 & 69.18 \\
   &   &   & \textbf{SFMP}$_{g128}$ & \textbf{5.31} & \textbf{8.36} & \textbf{78.07} \\ \cmidrule{2-7}
 & \multirow{5}{*}{30,531} & \multirow{5}{*}{3.25} 
 & GPTQ$_{w3g128}$ & 5.17 & 8.76 & 72.82 \\
 & & & AWQ$_{w3g128}$ & 4.78 & 8.57 & 75.18 \\
  & & & SliM-LLM$_{g128}$ & 4.74 & 8.52 & 77.41 \\
   & & & SqueezeLLM$_{0.45\%}$ & 4.71 & 8.48 & 75.16 \\
   &   &   & \textbf{SFMP}$_{g128}$ & \textbf{4.33} & \textbf{7.56} & \textbf{79.38} \\ \cmidrule{2-7}
 & \multirow{4}{*}{36,651} & \multirow{4}{*}{4.0} 
 & GPTQ$_{w4}$ & 4.58 & 8.42 & 74.88 \\
 & & & AWQ$_{w4}$ & 4.18 & 8.29 & 75.95 \\
& & & SqueezeLLM$_{w4}$ & 4.19 & 8.28 & 77.23 \\
  \multirow{-16}{*}{\textbf{\large 70B}} &   &   & \textbf{SFMP}$_{g128}$ & \textbf{3.37} & \textbf{7.01} & \textbf{80.47}  \\ \bottomrule \bottomrule
\end{tabular}
}
\caption{Evaluation of Llama3.1 8B/70B models on WikiText-2, C4 perplexity (PPL), and zero-shot tasks. Memory overhead from extra quantization parameters in GPTQ and AWQ at w3, w4 is omitted as it is negligible. SliM-LLM only supports group-wise quantization. Detailed zero-shot accuracy is provided in Table~\ref{tab:llama3.1_full_results_fixed_precison}.}
\label{tab:table_3}
\vspace{-0.4cm}
\end{table}

%% file: _analysis_ablation_study.tex
\vspace{-0.3cm}
\subsection{Analysis and Ablation Study}

More analysis and ablation study can be found in appendix~\ref{section:additinal_ablation_analysis}.

\textbf{Extra Metadata for blocks and reorder.}
SFMP requires storing additional metadata to index block-wise precision and activation reordering vectors. For a general weight matrix of size $[M, N]$, using a block size of $[512, 128]$, the matrix is partitioned into $\frac{M}{512} \times \frac{N}{128}$ blocks. Each block requires one $\text{int8}$ value to store its bit-width. In addition, we store two activation reordering vectors of sizes $[1, M]$ and $[1, N]$, both in $\text{fp16}$. Therefore, the average metadata overhead (in BPW) for a $L$-layers model can be written as: 
$$
\small
\frac{
\sum_{l=1}^{L} \Big(
    \tfrac{M_l}{512} \tfrac{N_l}{128}\,\text{sizeof(int8)}
    + (M_l + N_l)\,\text{sizeof(fp16)}
\Big)
}{
\sum_{l=1}^{L} M_l N_l
}
$$

In practice, taking LLaMA 3.1 8B as an example. The overhead is \textbf{0.006BPW}, which is negligible.
\begin{figure}[!t]
\vspace{-0.2cm}
\centering
\includegraphics[width=\linewidth]{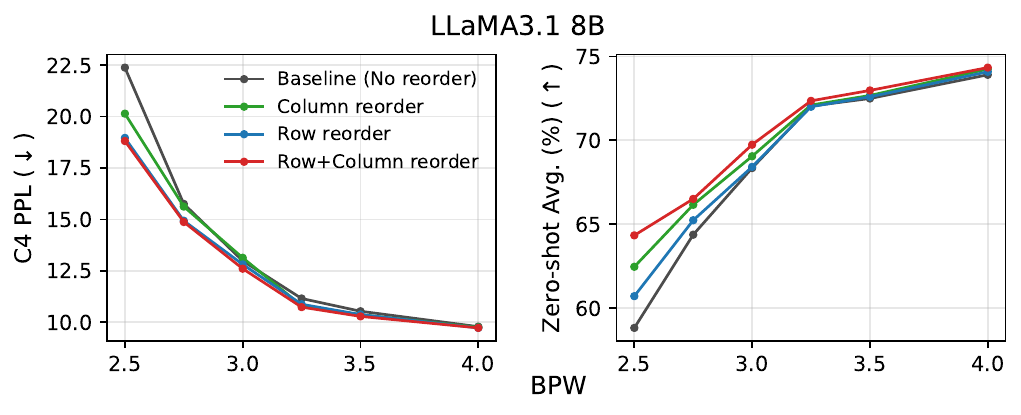} 
\vspace{-0.5cm}
\caption{Impact of row and column reordering across different average bits on model perplexity $(\downarrow)$ and zero-shot accuracy $(\uparrow)$.}
\vspace{-0.3cm}
\label{figure:impact_of_reorder}
\end{figure}

\begin{figure}[!t]
\centering
\includegraphics[width=\linewidth]{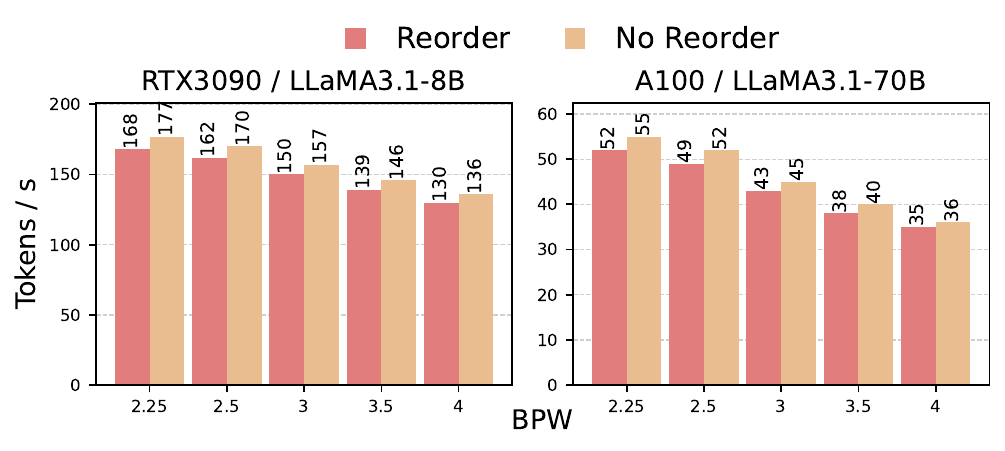} 
\vspace{-0.4cm}
\caption{Throughput (tokens / s) comparison for end-to-end generation of 128 tokens with and without reordering.}
\label{figure:overhead_of_reorder}
\vspace{-0.5cm}
\end{figure}

\textbf{Impact of row-column reordering.} 
We analyze the contribution of row-column reordering through an ablation study with four settings: no reorder, column reorder, row reorder, and combined row–column reorder. As shown in Fig.~\ref{figure:impact_of_reorder}, two key observations can be drawn: 1) Column reordering usually outperforms row reordering. This may be because it is better aligned with the activation-aware principle of AWQ, thereby more effectively protecting important input channels during quantization. 2) The performance gains from the row and column reordering  gradually diminish as the BPW increases. At BPW = 4, all four configurations achieve nearly identical accuracy. This convergence is due to AWQ already achieving near-lossless compression at 4-bit precision, which leaves little room for further improvement from reordering. Furthermore, Fig.~\ref{figure:overhead_of_reorder} shows the impact of reordering on end-to-end inference speed. Reordering incurs a modest slowdown of at most 5\%, with a smaller impact on larger models, as the increasing cost of GEMM amortizes the fixed overhead introduced by reordering.

\begin{figure}[!t]
\centering
\includegraphics[width=\linewidth]{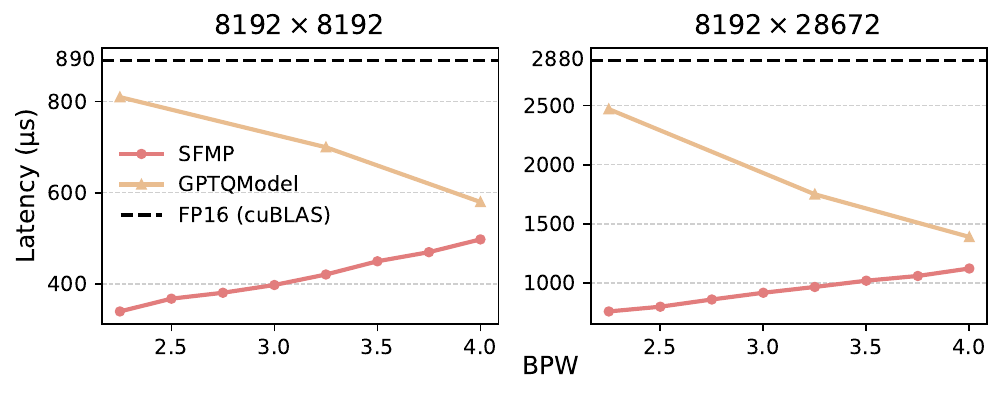} 
\caption{Latency comparison of our unified mixed-precision kernel , uniform quantization kernel from GPTQModel and cuBLAS FP16 kernel on A100.}
\vspace{-0.2cm}
\label{figure:kernel_evaluation}
\end{figure}

\textbf{Kernel evaluation.} Fig.~\ref{figure:kernel_evaluation} compares GEMV latency under three settings: FP16 (cuBLAS), the uniform quantization kernel from GPTQModel, and our unified kernel. For the uniform quantization kernel, we adopt the backend selected by GPTQModel’s automatic tuning. Specifically, at BPW=4, GPTQModel employs the state-of-the-art W4A16 kernel, Marlin~\cite{frantar2025marlin}. At BPW = 2 and 3, existing works provide limited kernel support and no widely adopted high-performance implementations, so GPTQModel uses a Triton-based implementation~\cite{tillet2019triton}. We select two representative GEMV operations from LLaMA3.1-70B: q\_proj and down\_proj. Across all shapes and bits, our kernel consistently achieves lower latency than both cuBLAS and GPTQModel. Notably, the latency of our kernel decreases approximately linearly with reducing BPW. This trend is attributed to eliminating weight dequantization overhead and leveraging LUT-based computation. In contrast, the latency of GPTQModel’s kernel increases with lower BPW due to the growing overhead of weight unpacking.


\textbf{Search cost.} Table~\ref{tab:search_time} compares the algorithmic cost of searching mixed-precision configurations for BitStack, AMQ, SliM-LLM and SFMP in terms of memory and time, evaluated on A100-80GB GPUs. BitStack incurs substantial overhead due to weight decomposition and block-wise sorting. AMQ reduces the search cost via proxy-based predictors and pruning, reducing the time to 44 hours for LLaMA3.1-70B. SliM-LLM can be executed with fewer GPUs, but configuring LLaMA3.1-70B still takes 8 hours. In contrast, SFMP directly assigns bit-widths by solving a special knapsack problem. Its main cost is estimating the diagonal values of the Fisher Information Matrix using a small calibration set, resulting in minimal algorithmic overhead.

\renewcommand{\arraystretch}{0.6}
\begin{table}[ht!]

\resizebox{\columnwidth}{!}{
\begin{tabular}{c||c|c||c|c}
\midrule
\textbf{ Model} & \multicolumn{2}{c||}{\textbf{ 8B}} & \multicolumn{2}{c}{\textbf{ 70B}} \\ \midrule
Parameter & \#GPU & Cost (h) & \#GPU & Cost (h)  \\ \midrule
SliM-LLM & 1 & 2 & 1 & 8  \\
BitStack & 1 & 12 & 4 & \textgreater{}300  \\
AMQ & 1 & 7 &  4 & 44 \\
\rowcolor{gray!25}SFMP & 1 & 0.15 & 4 & 0.50  \\ \bottomrule \bottomrule
\end{tabular}
}
\vspace{-0.2cm}
\caption{The search time on Llama 3.1 family of SFMP, BitStack, AMQ, SliM-LLM.}
\label{tab:search_time}
\end{table}

\begin{figure}[!t]
\centering
\includegraphics[width=\linewidth]{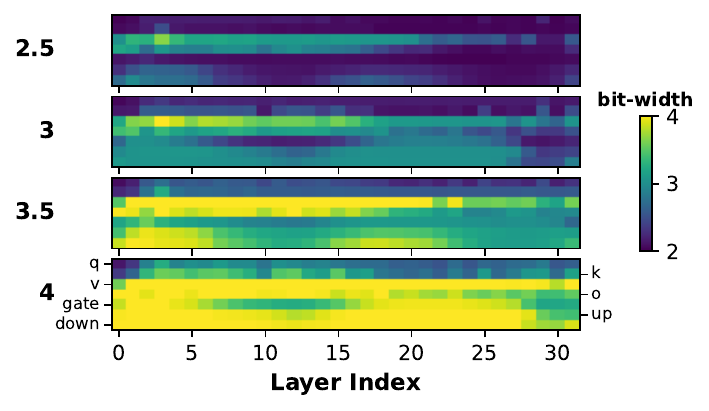} 
\vspace{-0.5cm}
\caption{Visualization of bit allocation over linear layers with different average bits at Llama3.1 8B. The numbers on the left indicate the average bit.}
\label{figure:bit_allocation}
\vspace{-0.4cm}
\end{figure}

\textbf{Bit allocation visualization.} Fig.~\ref{figure:bit_allocation} shows the bit allocation result on LLaMA3.1-8B. We provide detailed average bit-widths for each linear layer in Table~\ref{tab:layerwise_bits_compare}. It can be observed that the Value projection in self-attention consistently retains the higher bit-widths, followed by the Gate, Up, and Down layers, with Query and Key projections assigned the lower bit-widths. This patter is consistent with prior findings from AMQ \cite{lee2025amq}, validating the effectiveness of SFMP’s bit allocation scheme.

%% file: _conclusion.tex
\vspace{-0.2cm}
\section{Conclusion}
\vspace{-0.1cm}
SFMP formulates a fine-grained mixed-precision quantization problem based on a Taylor expansion of the model’s final output, simplifying the optimization objective and avoiding complex iterative search or black-box optimization. To maintain hardware friendliness, SFMP adopts a structured block-wise pattern, slightly sacrificing accuracy in exchange for regular memory layouts and efficient execution, and introduces a unified computation kernel. Overall, SFMP provides a practical solution for deploying large language models in resource-constrained environments.

%% file: _limitation.tex
\section*{Limitations}
\label{section:limitation_discussion}

Despite the effectiveness of the proposed method, several limitations remain and point to promising directions for future work. First, our current implementation and evaluation focus on GPU-based inference. Supporting additional hardware platforms such as CPUs, NPUs, and TPUs, would significantly broaden the applicability of our method. Second, current work focuses on weight-only quantization. Extending the optimization objective to include activation quantization would further improve inference efficiency, particularly in compute-bound scenarios. Third, as discussed in the Appendix~\ref{subsection:impact_of_block_size}, the group size plays a critical role in determining model accuracy under a fixed memory budget. However, existing mixed-precision quantization methods typically treat the group size as a fixed hyperparameter (e.g., 128), determined heuristically, and is independent of the bit allocation strategy. A promising future direction is to incorporate the group size into the mixed-precision optimization process and allow flexible, adaptive group sizes, which may further improve model performance.

%% file: _appendix.tex
\newpage
\appendix
\onecolumn

\section*{Appendix}

\addcontentsline{toc}{section}{Appendix}

\begin{center}
\textbf{Appendix Overview}
\end{center}

\noindent
\textbf{Appendix~\ref{section:additional_related_work}}: Additional Related Works 

\noindent
\textbf{Appendix~\ref{section:details_about_one-bit_gemm}}: Details about One-Bit Lut-Based GEMM

\noindent
\textbf{Appendix~\ref{section:derivation_of_global_salience}}: Fisher-Information-Based Analysis of Quantized Weight

\noindent
\textbf{Appendix~\ref{section:inference_speed_group_wise}}: Empirical Study on Inference Speed of Group-wise Mixed-Precision Methods

\noindent
\textbf{Appendix~\ref{section:empiriacl_analysis_weight_salience}}: Empirical Analysis of Fisher Diagonal Value Distribution 

\noindent
\textbf{Appendix~\ref{section:cuda_implementataion}}: Detailed CUDA Implementation

\noindent
\textbf{Appendix~\ref{section:additinal_ablation_analysis}}: Additional Ablation Analysis

\noindent
\textbf{Appendix~\ref{section:discussion_of_concurrent_work}}: Discussion of Related Concurrent Work

\noindent
\textbf{Appendix~\ref{section:SFMP_QAT}}: SFMP with Quantization-Aware Training

\noindent
\textbf{Appendix~\ref{section:autoregressive_result}}: Autoregressive Decoding Comparison Between SFMP and AMQ

\noindent
\textbf{Appendix~\ref{section:bit_allocation_visualizations}}: More Results of Bit Allocation Visualizations

\section{Additional Related Works}
\label{section:additional_related_work}

\subsection{Structured and Unstructured Quantization Format}
Structured quantization formats are generally more favorable for hardware execution.
For example, assigning a uniform integer precision to an entire linear layer enables regular memory access patterns and allows weights to be dequantized in a uniform manner, without introducing conditional branches or complex control flow. In large-scale models where such weight matrices appear extensively, this structured design is particularly advantageous for hardware acceleration \cite{frantar2022gptq,lin2024awq,chen2024efficientqat,lee2025amq,jang2025blockdialect}. In contrast, unstructured quantization formats typically offer finer granularity and greater flexibility. They allow the quantization precision to be adaptively adjusted according to the characteristics of individual weights or channels, and thus can achieve higher model accuracy under the same memory budget compared to structured quantization \cite{huang2024slim,jang2025blockdialect,li2023llm-mq,kim2023squeezellm,zhao2025ganq}.
However, this increased flexibility often comes at the cost of irregular memory access patterns and more complex dequantization procedures. As a result, unstructured quantization is generally less efficient in terms of inference latency and hardware utilization than structured quantization, especially on general-purpose accelerators.

\subsection{Layer-Wise Mixed-Precision}

Layer-wise mixed-precision quantization assigns different bit-widths to individual linear layers and typically formulates bit allocation as an integer programming or multi-objective optimization problem under a memory budget. However, this problem is NP-complete. For large-scale models, the search space becomes prohibitively large: LLaMA3.1 8B contains 224 linear layers, leading to a search space of $2^{224}$ even with only two candidate bit-widths, while LLaMA3.1 70B expands this space to $2^{560}$. To obtain acceptable solutions within a reasonable time, existing mixed-precision methods rely on heuristic strategies to reduce the search space. Most approaches \cite{cheng2025signroundv2,you2024shiftaddllm} adopt constrained formulations and solve the resulting integer programs using off-the-shelf solvers, whereas methods such as AMQ \cite{lee2025amq} cast bit allocation as a multi-objective optimization problem and employ genetic algorithms to approximate Pareto-optimal solutions.
\newpage 
\section{Details about One-Bit Lut-Based GEMM}
\label{section:details_about_one-bit_gemm}

\begin{figure}[bt!]
    \centering
    \includegraphics[width=\textwidth]{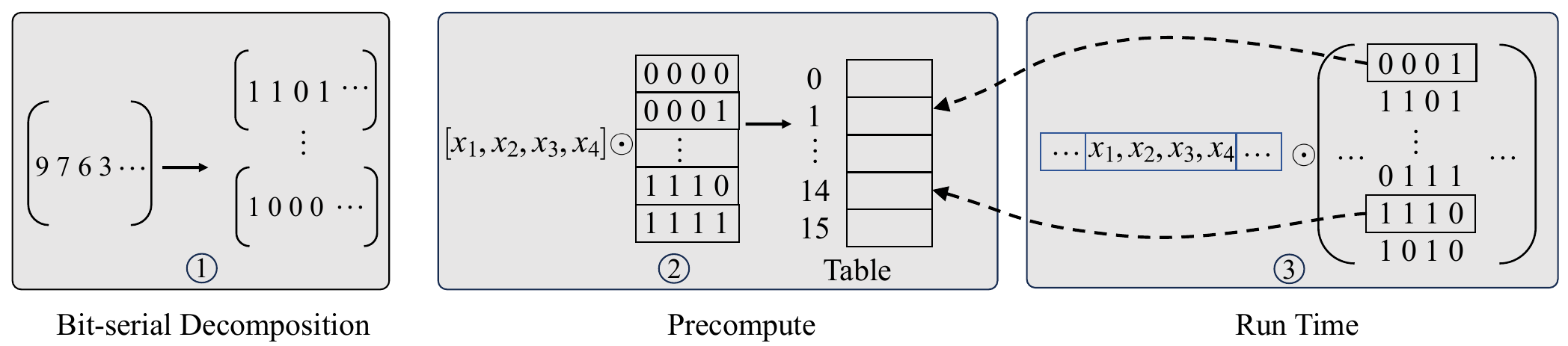}
    \caption{Detailed computation procedure of one-bit Lut-based GEMM.}
    \label{figure:one-bit-gemm}
\end{figure}

Fig.~\ref{figure:one-bit-gemm} illustrates the detailed computational procedure of one-bit Lut-based GEMM. First, a $q$-bit quantized weight matrix $\mathrm{W}_{int} \in \mathbb{Z}^{m \times n}$ is decomposed into $q$ one-bit matrices $\{W_0, W_1, ..., W_{q-1}\}$, where $W_i \in \{ {0,1\}}^{m\times n}$, representing the respective bit planes of the original weights. For example, for integer values (9, 7, 6, 3) with binary representations (1001, 0111, 0110, 0011), the vector for the lowest bit is (1, 1, 0, 1), and the vector for the highest bit is (1, 0, 0, 0). This decomposition is performed offline, incurring no runtime overhead. During inference, for an activation vector of the group size $g$, the operator precomputes the dot products between this activation vector and all $2^g$ possible combinations of one-bit weights, storing the results in the LUT. Thus, the original matrix computation requiring high-precision multiply-accumulate operations is simplified into highly efficient table lookups followed by summation. As shown in the Eq.~\ref{equation:one-bit-gemm}, the matrix multiplication between the activation $X$ and the original quantized weight $W_{int}$ can be transformed into a sum of multiple one-bit GEMM operations:
\begin{equation}
\label{equation:one-bit-gemm}
X \times W_{int} = X \times \left( \sum_{i=0}^{q-1} 2^i W_i \right)
= \sum_{i=0}^{q-1} 2^i \, X \times W_i, \ \   W_i\in\{0,1\}^{m\times n} .
\end{equation}
To reduce the table size and accelerate table lookup, a commonly used technique is mirror storage. Typically, dequantized weight $\hat W$ can be written in the following form:
\begin{equation}
\hat{\mathrm{W}}=\sum_{i=0}^{q-1}2^{i}s W_i+z, \ \ W_i\in\{0,1\}^{m\times n},
\end{equation}
where $s\in \mathbb{R}$ denotes the scale and $z \in \mathbb{R}$ denotes the zero-point. we apply a simple linear transformation by setting $\hat s = \frac 1 2 s$, $\hat W_i = 2W_i - 1 $, $\hat z = z +  \frac 1 2 \sum_{i=0}^{q-1}2^is$. After that, the dequantized weight $\hat W$ can be rewritten as:
\begin{equation}
\mathrm{\hat W} = \sum_{i=0}^{q-1} 2^{i} \hat s \hat W_i + \hat z,\ \ \hat W_i \in \{ -1,1\}^{m\times n}.
\end{equation}
Under this transformation, for example, with an input activation combinations $[x_1,x_2,x_3,x_4]$, the output of the dot product has 16 possible outcomes, ranging from $(-x_1-x_2-x_3-x_4,\ ...,\ x_1+x_2+x_3+x_4)$. When storing the lookup table, we only need to store half of the possible results, as the remaining half can be obtained by negating the stored values. This table compression method is lossless, fully preserving model inference accuracy while also reducing memory usage by half and accelerating table access.

The one-bit Lut-based GEMM has been demonstrated to offer high computational efficiency and energy efficiency \cite{park2025figlut,Weitmac}. FIGLUT \cite{park2025figlut} optimized the table structure for GPU architectures to avoid bank conflicts, while T-MAC \cite{Weitmac} leveraged CPU vectorized lookup instructions (AVX2/NEON) to enable efficient LUT operations on CPUs.

\newpage 
\section{Fisher-Information-Based Analysis of Quantized Weight}
\label{section:derivation_of_global_salience}

The objective of quantization is to approximate the original full-precision weight $W$ with their quantized representation $W'$, while minimizing the degradation of the final task loss. To obtain a reliable measure, we aim to characterize how sensitive the \emph{final loss} is to perturbations of individual weights.

Let $\mathcal{L}(W)$ denote the original output of the model with weights $W$.
When the weights are perturbed by quantized weights $W'$, the change in loss can be approximated by a second-order Taylor expansion:
\begin{equation}
\mathcal{L}(W) - \mathcal{L}(W') \approx g\top (W-W')
  + \frac{1}{2} (W-W')^\top H (W-W'),
\end{equation}
where $g = \nabla_W \mathcal{L}(W)$ and 
$H = \mathbb{E}\!\left[\frac{\partial^2 \mathcal{L}(W)}{\partial W^2}\right]$
are gradient and the Hessian of the loss.

Since the model is assumed to be well-trained, the gradient term vanishes in expectation,
$\nabla_W \mathcal{L}(W) \approx 0$,
and the dominant contribution to the loss increase induced by quantization comes from the second-order term:
\begin{equation}
\Delta \mathcal{L}
\approx
\frac{1}{2} (W-W')^\top H (W-W').
\end{equation}
This expression reveals that the effect of weight perturbations on the final loss is governed by the curvature of the loss landscape. Perturbations along directions with large curvature lead to disproportionately larger loss increases.

Direct computation of the Hessian is infeasible for large-scale models.
Following prior work \cite{kim2023squeezellm}, we approximate the Hessian using the Fisher Information Matrix \cite{ly2017fisherinformation}:
\begin{equation}
H \simeq \mathcal{F}
=
\mathbb{E}_{(x,y)\sim D}
\left[
\nabla_W \log p(y|x;W)\,
\nabla_W \log p(y|x;W)^\top
\right],
\end{equation}
which can be efficiently estimated using gradients computed over a sample dataset $D$.
This approximation is well-motivated for maximum-likelihood objectives and has been widely adopted in previous works~\cite{frantar2022gptq,tang2022mixed,kim2023squeezellm}.

To further reduce computational complexity, we assume that cross-weight interactions are negligible and approximate the Fisher matrix by its diagonal:
\begin{equation}
\mathcal{F} \approx \mathrm{diag}(\mathcal{F}_{11}, \dots, \mathcal{F}_{NN}).
\end{equation}
Under this diagonal approximation, the expected increase in loss induced by parameter perturbations decomposes into a sum of independent per-weight contributions:
\begin{equation}
\Delta \mathcal{L}
\approx
\frac{1}{2}
\sum_{i=1}^{N}
\mathcal{F}_{ii}\, (W_i-W_i')^2.
\end{equation}

\newpage 
\section{Empirical Study on Inference Speed of Group-wise Mixed-Precision Methods}
\label{section:inference_speed_group_wise}

\begin{figure*}[!t]
\centering
\includegraphics[width=\linewidth]{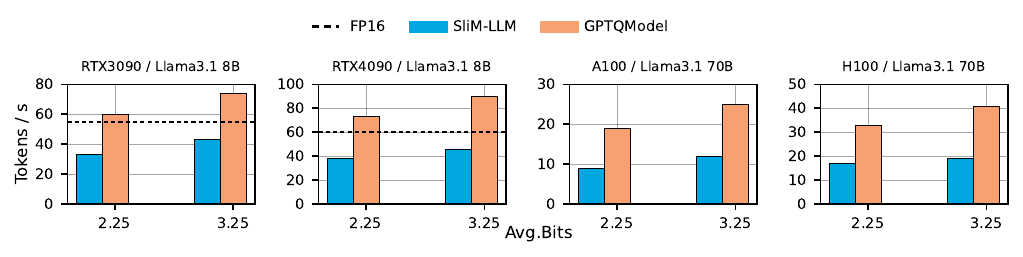} 
\caption{Inference throughput (tokens / s) comparison between SliM-LLM and GPTQ when generating 128 tokens with batch size 1. BF16 inference of LLaMA3.1 70B is not feasible on single A100 and H100 due to memory constraints. BPW denotes \textit{``bits per weight"}}
\label{figure:empirical_study_slim}
\end{figure*}

We present an empirical study that compares the inference throughput of the group-wise mixed-precision method SliM-LLM \cite{huang2024slim} and the uniform quantization method GPTQ \cite{frantar2022gptq}. For SliM-LLM, we use the official released code, while GPTQ is evaluated using GPTQModel\footnote{\url{https://github.com/ModelCloud/GPTQModel}}. As shown in Fig.~\ref{figure:empirical_study_slim}, at the same BPW (bits per weight), SliM-LLM exhibits a substantial reduction in inference throughput compared to GPTQ, with a slowdown of up to 50\%. The result indicates that group-wise mixed-precision quantization introduces significant hardware inefficiencies, leading to a reduced inference speed.

\section{Empirical Analysis of Fisher Diagonal Value Distribution}
\label{section:empiriacl_analysis_weight_salience}
\begin{figure*}[!t]
\centering
\includegraphics[width=\textwidth]{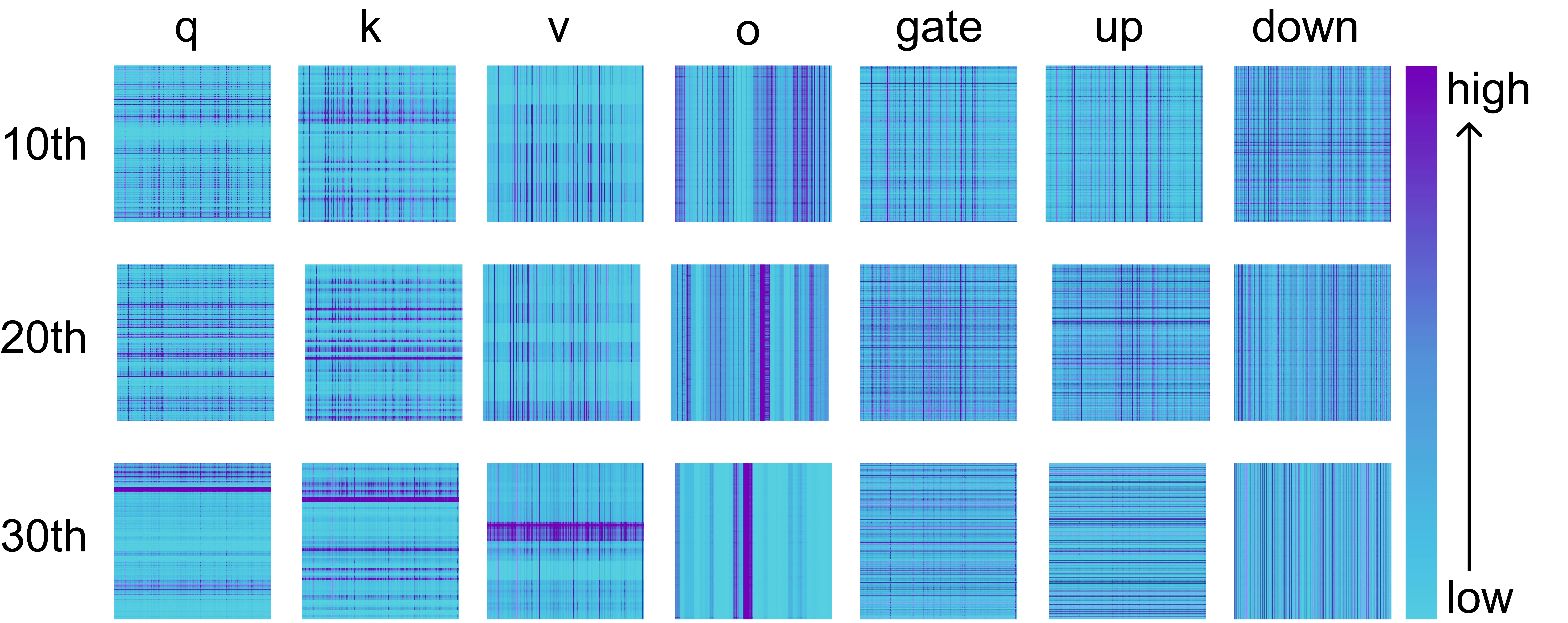} 
\caption{Weight salience distribution in the $10^{th}$, $20^{th}$, $30^{th}$ layers of LLaMA3.1 8B}
\label{figure:salience_distribution}
\end{figure*}

Fig.~\ref{figure:salience_distribution} illustrates the distribution of Fisher diagonal values in LLaMA3.1 8B. It can be observed that the values tend to concentrate along rows or columns of the weight matrix, rather than forming spatially contiguous blocks.

\newpage

\section{Detailed CUDA Implementation}
\label{section:cuda_implementataion}

\begin{figure*}[bt!]
    \centering
    \includegraphics[width=\textwidth]{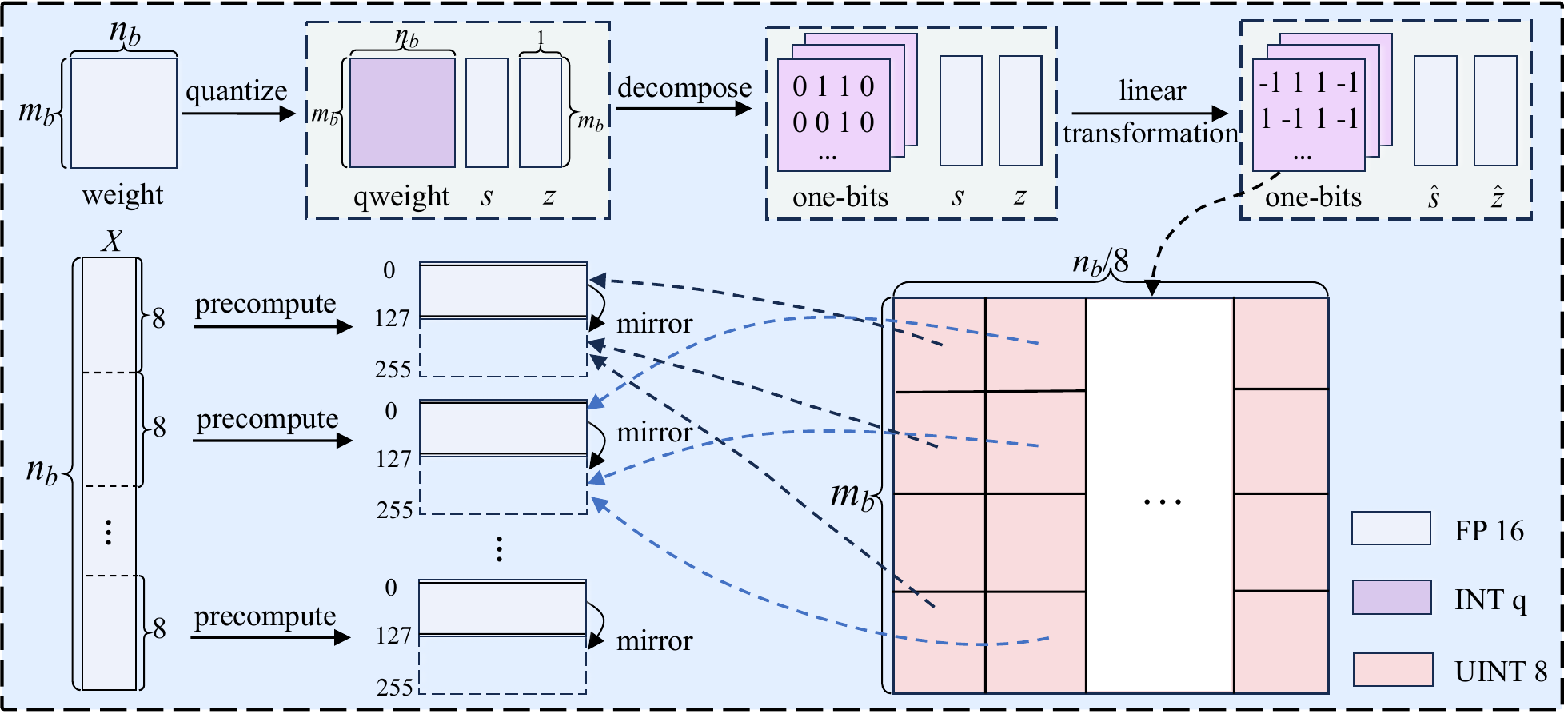}
    \caption{CUDA implementation.}
    \label{figure:cuda_implementation}
    \vspace{-0.5cm}
\end{figure*}

Fig.~\ref{figure:cuda_implementation} shows our CUDA implementation: a quantized matrix-vector multiplication. After applying a quantization algorithm (e.g., AWQ), the integer weights within each block are decomposed into multiple one-bit components. We then apply an equivalent linear transformation, as described in Appendix~\ref{section:details_about_one-bit_gemm}, which facilitates the subsequent construction of lookup tables. The transformed weights are finally packed along the $n_b$ dimension into \texttt{uint8} values.

During matrix–vector multiplication, input activations are grouped into 8-element vectors, and the corresponding dot products for all $2^8 = 256$ possible activation combinations are precomputed and stored in a lookup table. Owing to the applied linear transformation, only 128 entries need to be explicitly constructed, while the remaining entries can be obtained via mirror.

\begin{figure*}[bt!]
    \centering
    \includegraphics[width=\textwidth]{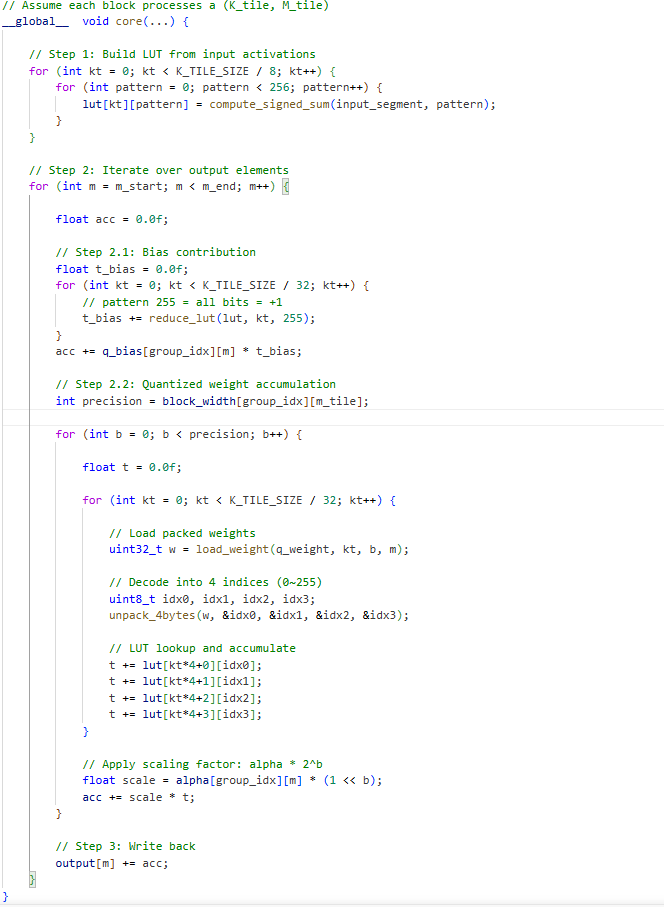}
    \caption{Pseudo code for CUDA.}
    \label{figure:pseudo_cuda}
    \vspace{-0.5cm}
\end{figure*}

Once the lookup table is constructed, the quantized weights, stored as packed \texttt{uint8} values, are used to index the table and perform accumulation. The unified LUT kernel constructs a shared-memory lookup table on-the-fly. Each thread block, responsible for a tile of size $[M_{\text{tile}},K_{\text{tile}}]$,builds a LUT of size $[K_{\text{tile}}/8,256]$. With $K_{\text{tile}}=64$, this corresponds to $8\times256$ entries, where 256 enumerates all $2^8$ sign combinations over 8 input elements. Each LUT entry is built from 8 inputs and expanded via lightweight accumulation, mapping each 8-bit index to a partial dot product. Packed weights then index the LUT and are accumulated across bit-planes with quantization scales. The LUT is constructed once per thread block and reused across all $M_{\text{tile}}=512$ output channels. The memory overhead of shared memory $8\times256\times2$ bytes, about \textbf{4KB} per thread block. Fig.~\ref{figure:pseudo_cuda} shows the pseudo code for CUDA.

\section{Additional Ablation Analysis}
\label{section:additinal_ablation_analysis}
\subsection{Impact of Block Size}
\label{subsection:impact_of_block_size}

We study the impact of block size $(m_b, n_b)$ on model accuracy.

\subsubsection{Effect of $m_b$.} 
As shown in Table~\ref{table:impact_of_bm}, decreasing $m_b$ consistently improves model accuracy. This behavior is expected, as a smaller $m_b$ corresponds to a finer granularity of block-wise mixed-precision. However, when $m_b < 512$, further reducing $m_b$ only yields marginal accuracy gains. Therefore, in practice, considering GPU hardware characteristics such as warp size and thread scheduling, we typically choose $m_b \in \{256, 512\}$ to achieve a good balance between accuracy and efficiency.

\begin{table*}[ht]
\centering

\resizebox{\textwidth}{!}{
\begin{tabular}{c|c|ccccc}
\toprule
\textbf{Model} & \textbf{BPW} & $m_b{=}64$ & $m_b{=}128$ & $m_b{=}256$ & $m_b{=}512$ & $m_b{=}1024$ \\
\midrule
\multirow{3}{*}{LLaMA3.1 8B}

 & 2.50 & \textbf{(13.60, 66.15)} & (13.64, 66.08) &   (13.66, 66.10)&  (13.68, 65.97)  & (14.93, 63.68) \\ \cmidrule{2-7}
 & 3.00 & \textbf{(8.59, 69.85)}  & (8.61, 69.78)  & (8.62, 69.47) & (8.65, 69.92)    & (9.57, 68.96) \\ \cmidrule{2-7}

 & 3.50 & \textbf{(7.28, 73.60)}  & (7.29, 73.68)  & (7.28, 73.59) & (7.30, 73.43)    & (7.48, 72.78) \\ 
\midrule
\multirow{3}{*}{LLaMA3.1 70B}

 & 2.50 & \textbf{(7.21, 74.80)} & (7.22, 74.72) & (7.24, 74.65) & (7.24, 74.60)  & (7.31, 74.02) \\ \cmidrule{2-7}
 & 3.00 & \textbf{(5.29, 78.20)} & (5.31, 78.02) & (5.30, 78.14) & (5.31, 78.07)  & (5.40, 77.85) \\
 \cmidrule{2-7}
 & 3.50 & \textbf{(4.00, 80.02)} & (4.00, 79.94) & (4.00, 80.15) & (4.00, 80.07)  & (4.00, 79.66) \\
 
\bottomrule
\end{tabular}
}
\caption{Ablation study of block size $m_b$ under different BPWs.
Each entry reports (WikiText2 perplexity ($\downarrow$), Zero-shot average accuracy (\%) ($\uparrow$)), with $n_b$ fixed to 128.}
\label{table:impact_of_bm}
\end{table*}

\subsubsection{Effect of $n_b$.} 
In contrast to $m_b$, the choice of $n_b$ exhibits a more pronounced and non-monotonic effect on accuracy. In our quantization scheme, each block applies group quantization with a group size of $n_b$. The parameter $n_b$ directly controls both quantization granularity and the storage overhead of quantization parameters.

Under a fixed memory budget, a smaller $n_b$ leads to higher overhead for storing scale and zero-point parameters. For example, assuming that the scales and zero-points are stored in BF16, when $n_b = 128$, they require an average of 0.25 bits per weight. This overhead increases to 0.5 bits when $n_b = 64$, and decreases to 0.125 bits when $n_b = 256$. As shown in Table~\ref{table:impact_of_bn}, if $n_b$ is too small, excessive budget is consumed by quantization parameters, leaving insufficient bit-width for the weights themselves and degrading model accuracy. Conversely, if the group size is too large, the value distribution within a group may become highly heterogeneous, and uniform quantization introduces large quantization errors, which also harms performance.

Consequently, $n_b$ is neither \textbf{``the smaller the better''} nor \textbf{``the larger the better,''}. Notably, prior mixed-precision methods typically fix the group size (e.g., $n_b = 128$) and overlook its impact on the accuracy--budget trade-off. Our ablation analysis demonstrate that careful selection of group size is essential for achieving optimal accuracy under fixed memory budget. We leave adaptive group size selection under a fixed memory budget as an interesting direction for future work.

\begin{table}[ht]
\centering

\resizebox{\textwidth}{!}{%
\begin{tabular}{c|c|cccc}
\toprule
\textbf{Model} & \textbf{BPW} & $n_b{=}64$ & $n_b{=}128$ & $n_b{=}256$ & $n_b{=}512$ \\
\midrule
\multirow{6}{*}{LLaMA3.1 8B}
 & 2.25 & (4894, 36.79) & (2520, 37.42) & \textbf{(28.61, 57.69)} & (43.56, 57.39) \\ 
 \cmidrule{2-6}
 & 2.50 & (737, 40.90)  & (13.68, 65.97) & \textbf{(13.32, 66.54)} & (17.93, 62.94) \\
  \cmidrule{2-6}
 & 3.00 & (10.12, 68.40) & \textbf{(8.65, 69.92)}  & (8.69, 69.74)  & (9.53, 68.58) \\
  \cmidrule{2-6}
 & 3.25 & (8.83, 69.99)  & (7.78, 72.47)  & \textbf{(7.60, 72.98)}  & (7.94, 71.42) \\
  \cmidrule{2-6}
 & 3.50 & (7.98, 71.76)  & \textbf{(7.30, 73.43)}  & (7.51, 72.26)  & (7.59, 72.15) \\
\midrule
\multirow{6}{*}{LLaMA3.1 70B}
 & 2.25 & (2746, 45.11) & (1482, 47.52) & \textbf{(8.17, 72.65)} & (11.26, 68.43) \\
 \cmidrule{2-6}
 & 2.50 & (235, 57.36)  & (7.24, 74.60) & \textbf{(7.13, 75.12)} & (7.64, 73.85) \\
 \cmidrule{2-6}
 & 3.00 & (5.47, 77.25) & \textbf{(5.31, 78.07)} & (5.36, 77.94) & (5.40, 77.80) \\
 \cmidrule{2-6}
 & 3.25 & (4.98, 76.13) & (4.60, 76.71) & \textbf{(4.33, 79.38)} & (4.28, 79.56) \\
 \cmidrule{2-6}
 & 3.50 & (4.13, 79.49) & \textbf{(4.00, 80.07)} & (4.02, 80.25) & (4.05, 80.48) \\
\bottomrule
\end{tabular}
}
\caption{Ablation study of block size $n_b$.
Each entry reports (WikiText2 perplexity ($\downarrow$), Zero-shot average accuracy (\%) ($\uparrow$)), with $m_b$ fixed to 512.}
\label{table:impact_of_bn}
\end{table}

\subsection{Impact of Sample Size for Fisher Estimation}
Table~\ref{table:impact_of_sample_size} reports the impact of the sample size used for Fisher information estimation on model performance. As shown in the table, increasing the sample size beyond 512 leads to only marginal improvements in model performance across different BPWs. Based on this observation, we adopt a sample size of 1K throughout our work as a reasonable trade-off between estimation accuracy and computational cost. Notably, even with a sample size of 128, our method is still able to achieve competitive performance, indicating a certain degree of robustness to imperfect Fisher estimation. In addition, model performance under the lower BPW exhibits higher sensitivity to the sample size. This trend further highlights the importance of accurately identifying salient weights when performing low-bit quantization.

\begin{table*}[ht]
\centering

\resizebox{\textwidth}{!}{%
\begin{tabular}{c|c|ccccc}
\toprule
\textbf{Model} & \textbf{BPW} & 128 & 256 & 512 & \textbf{1024} & 2048 \\
\midrule
\multirow{4}{*}{\textbf{LLaMA3.1 8B}} & 2.5 & (13.89,65.02) & (13.81,65.28) & (13.73,65.11) & (13.68,65.97) & (13.66,66.07) \\
\cmidrule{2-7}
& 3  &(8.80,69.34)  & (8.80,69.45)  & (8.78,69.60)  & (8.65,69.92)  & (8.63,69.98)  \\
\cmidrule{2-7}
& 3.5 & (7.40,72.89)  & (7.39,72.93)  & (7.35,72.78)  & (7.30,73.43)  & (7.30,73.56)  \\
\cmidrule{2-7}
& 4 & (6.86,73.72)  & (6.86,73.88)  & (6.86,73.91)  & (6.84,74.15)  & (6.84,74.28)  \\

\midrule
\multirow{4}{*}{\textbf{LLaMA3.1 70B}} & 2.5 & (7.30,74.32)  & (7.26,74.54)  & (7.26,74.67)  & (7.24,74.60)  & (7.23,74.71)  \\
\cmidrule{2-7}
& 3 & (5.38,77.82)  & (5.33,78.10)  & (5.33,77.92)  & (5.31,78.07)  & (5.30,78.19)  \\
\cmidrule{2-7}
& 3.5 & (4.10,80.02)  & (4.05,79.85)  & (4.03,79.93)  & (4.00,80.07)  & (4.00,80.13)  \\
\cmidrule{2-7}
& 4 &(3.43,80.50)  & (3.44,80.29)  & (3.38,80.02)  & (3.37,80.47)  & (3.38,80.23)  \\
\bottomrule
\end{tabular}
}
\caption{Impact of sample size for Fisher estimation on model performance. Each entry reports (WikiText2 perplexity ($\downarrow$), Zero-shot average accuracy (\%) ($\uparrow$))}
\label{table:impact_of_sample_size}
\end{table*}

\subsection{Impact of Calibration Sets}
Table~\ref{table:calib_set_impact} shows that our method remains robust across different calibration sets, with negligible fluctuations in the overall average zero-shot accuracy.

\begin{table}[ht]
\centering

\begin{tabular}{c|c|cc}
\toprule
\multirow{2}{*}{\textbf{Model}} & \multirow{2}{*}{\textbf{BPW}} & \multicolumn{2}{c}{\textbf{Dataset}} \\
\cmidrule{3-4}
&  & \textbf{WikiText2} & \textbf{C4} \\
\midrule
\multirow{4}{*}{\textbf{LLaMA3.1 8B}} & 2.5 & 65.78 & 65.97 \\
\cmidrule{2-4}
& 3 & 69.98 & 69.92 \\
\cmidrule{2-4}
& 3.5 & 73.21 & 73.43 \\
\cmidrule{2-4}
& 4 & 74.20 & 74.15 \\

\midrule
\multirow{4}{*}{\textbf{LLaMA3.1 70B}} & 2.5 & 74.42 & 74.60 \\
\cmidrule{2-4}
& 3 & 78.16 & 78.07 \\
\cmidrule{2-4}
& 3.5 & 79.86 & 80.07 \\
\cmidrule{2-4}
& 4 & 80.45 & 80.47 \\
\bottomrule
\end{tabular}
\caption{Impact of calibration dataset on model performance. Zero-shot average accuracy (\%) ($\uparrow$) is reported.}
\label{table:calib_set_impact}
\end{table}

\subsection{Comparison with LUT-based GEMM Family}

We further compare SFMP with a representative LUT-based GEMM baseline, AnyBCQ~\cite{park2025anybcq} on A100. As both approaches are based on LUT-based computation, this comparison provides a more direct assessment of kernel efficiency within the same optimization family. As shown in table~\ref{table:lut_ablation}, SFMP achieves comparable kernel latency to AnyBCQ across different bit-widths and matrix sizes, with the performance gap consistently within 3\%. This small latency gap is mainly attributed to the additional runtime computation in SFMP required to determine the actual memory offsets of blocks in the flattened weight layout. The experiment indicates that SFMP preserves the efficiency of LUT-based GEMM implementations while introducing a more flexible block-wise quantization scheme.

\begin{table}[ht]
\centering

\begin{tabular}{c|ccc}
\toprule
\textbf{Bits} & \textbf{(M, K)} & \textbf{AnyBCQ (ms)} & \textbf{SFMP (ms)} \\
\midrule
\multirow{2}{*}{2}
& (8192,8192) & 332 & 340 \\
\cmidrule{2-4}
& (28672,8192) & 755 & 760 \\
\midrule
\multirow{2}{*}{3}
& (8192,8192) & 410 & 421 \\
\cmidrule{2-4}
& (28672,8192) & 942 & 967 \\
\midrule
\multirow{2}{*}{4}
& (8192,8192) & 478 & 495 \\
\cmidrule{2-4}
& (28672,8192) & 1076 & 1102 \\
\bottomrule
\end{tabular}
\caption{Kernel latency comparison (ms) with the LUT-based GEMM baseline (AnyBCQ).}
\label{table:lut_ablation}
\end{table}

\subsection{Analysis of Step Size $\Delta$}
\label{section:analysis_of_search_step}
We conduct an ablation study on the one-dimensional grid search step size $\Delta$ to evaluate its impact on model performance, with results summarized in table~\ref{table:step_ablation}. We observe that a step size $\Delta = 0.01$ already achieves strong performance. Further reducing the step size below 0.01 yields only marginal improvements. Therefore, we adopt $\Delta = 0.01$ in our method.

\begin{table}[ht]
\centering

\begin{tabular}{c|c|cccc}
\toprule
\multirow{2}{*}{\textbf{Model}} & \multirow{2}{*}{\textbf{BPW}} & \multicolumn{4}{c}{$\Delta$} \\ 
\cmidrule{3-6}
 & & 0.005 & \textbf{0.01} & 0.02  & 0.04 \\
\midrule
\multirow{6}{*}{\textbf{LLaMA3.1 8B}} 
& 2.35 & (24.55, 60.91) & (24.57, 60.80) & (24.68, 60.51) & (24.82, 60.51) \\
\cmidrule{2-6}
& 2.5  & (13.68, 65.92) & (13.68, 65.97) & (13.75, 65.23) & (13.94, 64.89) \\
\cmidrule{2-6}
& 3    & (8.64, 69.97)  & (8.65, 69.92)  & (8.67, 69.75)  & (8.74, 69.26) \\
\cmidrule{2-6}
& 3.25 & (7.78, 72.56)  & (7.78, 72.47)  & (7.80, 72.10)  & (7.85, 71.78) \\
\cmidrule{2-6}
& 3.5  & (7.30, 73.56)  & (7.30, 73.43)  & (7.31, 73.34)  & (7.33, 73.12) \\
\cmidrule{2-6}
& 4    & (6.84, 74.22)  & (6.84, 74.15)  & (6.84, 74.08)  & (6.84, 73.82) \\
\bottomrule
\end{tabular}
\caption{Ablation on grid search step size $\Delta$. Each entry reports (WikiText2 perplexity ($\downarrow$), Zero-shot accuracy (\%) ($\uparrow$)). Result shows that $\Delta=0.01$ achieves strong performance, while smaller step sizes bring negligible gains.}
\label{table:step_ablation}
\end{table}

\subsection{Analysis of Candidate Bit-Width $\mathcal{B}$}
\label{section:analysis_of_candidate_bitwidth}

We conduct an ablation study on the number of candidate bit-widths in $\mathcal{B}$, with results reported in table~\ref{table:bitwidth_ablation}. Across different BPW settings, expanding $|\mathcal{B}|$ from 2 to 3 consistently improves both perplexity and zero-shot accuracy. For instance, at BPW = 2.5, zero-shot accuracy increases from 64.34 to 65.97. Similar gains are observed across other configurations. However, further increasing $|\mathcal{B}|$ to 4 yields only marginal improvements, and in some cases the performance remains nearly unchanged (e.g., BPW = 3.5 and 4), indicating diminishing returns as the number of candidate bit-widths grows. So $\left|\mathcal{B}\right| = 3$ provides a favorable trade-off between model performance and optimization complexity.

\begin{table}[ht]
\centering

\begin{tabular}{c|c|c|c|c}
\toprule
\multirow{2}{*}{\textbf{Model}} & \multirow{2}{*}{\textbf{BPW}} &  \multicolumn{3}{c}{\textbf{$|\mathcal{B}|$}}  \\ 
\cmidrule{3-5}
 & & 2 & \textbf{3} & 4 \\
\midrule
\multirow{6}{*}{\textbf{LLaMA3.1 8B}} 
& 2.35 & (27.13, 58.06) & (24.57, 60.80) & (24.53, 60.74) \\
\cmidrule{2-5}
& 2.5  & (14.49, 64.34) & (13.68, 65.97) & (13.68, 66.12) \\
\cmidrule{2-5}
& 3    & (9.51, 68.74)  & (8.65, 69.92)  & (8.60, 69.77) \\
\cmidrule{2-5}
& 3.25 & (8.08, 71.32)  & (7.78, 72.47)  & (7.76, 72.16) \\
\cmidrule{2-5}
& 3.5  & (7.59, 72.23)  & (7.30, 73.43)  & (7.30, 73.65) \\
\cmidrule{2-5}
& 4    & (6.88, 73.87)  & (6.84, 74.15)  & (6.84, 74.22) \\
\bottomrule
\end{tabular}
\caption{Ablation on the number of candidate bit-widths $|\mathcal{B}|$. Each entry reports (WikiText2 perplexity ($\downarrow$), Zero-shot accuracy (\%) ($\uparrow$)). Results show that $\mathcal{B}=3$ achieves strong performance.}
\label{table:bitwidth_ablation}
\end{table}

\subsection{Comparison with Additional 4-bit Inference Baselines}
Although TensorRT-LLM~\cite{tensorrt_llm} and ExLlamaV1~\cite{exllama} do not support 2-bit or 3-bit model inference, they provide specialized optimizations for 4-bit inference, particularly under the batchsize=1 setting. In the Table~\ref{tab:4bit_baselines}, we further report the end-to-end inference speed comparison between SFMP and these inference backends at BPW=4. It can be observed that SFMP still demonstrates outstanding performance.

\begin{table}[t]
\centering

\begin{tabular}{c|c|cc}
\toprule
Hardware & Model  & Backend & Throughput (tokens/s) \\
\midrule
\multirow{4}{*}{RTX4090} 
& \multirow{4}{*}{LLaMA3.1 8B} 
& FP16         & 55  \\
& &  ExLlamaV1   & 114 \\
& & TensorRT-LLM & 143 \\
& & \textbf{SFMP} & \textbf{150} \\
\midrule
\multirow{4}{*}{A100} 
& \multirow{4}{*}{LLaMA3.1 70B} 
& FP16         & -  \\
& &  ExLlamaV1   & 28 \\
& & TensorRT-LLM & 32 \\
& & \textbf{SFMP} & \textbf{35} \\
\bottomrule
\end{tabular}
\caption{Comparison with additional 4-bit inference baselines when generating a sequence length of 128 with batchsize of 1.}
\label{tab:4bit_baselines}
\end{table}

\section{Discussion of Related Concurrent Work}
\label{section:discussion_of_concurrent_work}
Recently, a concurrent work, ScaleBits~\cite{li2026scalebits}, also explores block-wise quantization. However, our method is developed independently and differs in several key aspects. First, our approach adopts a closed-form, search-free formulation for bit allocation, whereas ScaleBits relies on an iterative search-based optimization procedure. Second, we design a unified LUT-based GEMM kernel that supports matrix multiplication between activations and weight matrices composed of arbitrary mixed bit-width blocks. In contrast, ScaleBits still relies on a dequantization-based computation kernel. As the code of Scalebits is not open-sourced, we will add experimental comparisons in the future.

\newpage 
\section{SFMP with Quantization-Aware Training}
\label{section:SFMP_QAT}
The block-wise mixed-precision format of SFMP is orthogonal to most quantization tuning techniques. As shown in Table~\ref{table:sfmp_qat}, by integrating SFMP with EfficientQAT \cite{chen2024efficientqat}, an advanced quantization-aware training (QAT) method, we further improve model accuracy.

\renewcommand{\arraystretch}{0.8}
\begin{table*}[ht]

\setlength{\tabcolsep}{1mm}
\resizebox{\textwidth}{!}{%
\begin{tabular}{c|c|c||cc||ccccccc}
\midrule
\textbf{\large Model}  & \textbf{\large BPW} &  \textbf{\large Method} &  \textbf{\large Wiki2($\downarrow$)} &  \textbf{\large C4($\downarrow$)} & \textbf{\large HellaS.($\uparrow$)} &  \textbf{\large WinoG.($\uparrow$)} &  \textbf{\large ARC-e($\uparrow$)} &  \textbf{\large ARC-c($\uparrow$)} &  \textbf{\large PIQA($\uparrow$)} & \textbf{\large BoolQ($\uparrow$)} & \textbf{\large Avg.($\uparrow$)} \\ \midrule
\multirow{7}{*}{\large \textbf{L3.1 8B}}  & 16 & BF16 & 6.15 & 8.89 & 78.99 & 72.93 & 81.19 & 53.41 & 81.39 & 82.15 & 75.01  \\ 
\cmidrule{2-12}
  & \multirow{2}{*}{2.25} & EfficientQAT$_{w2g128}$ & 13.20	&14.86	&		64.96&	64.64&	63.97	&37.71&	75.03&	71.77&	63.01\\
 &  & SFMP++$_{g256}$ & \textbf{10.89}&	\textbf{13.32}		&	\textbf{69.29}&	\textbf{67.17}&	\textbf{69.44}	& \textbf{40.61}&	\textbf{76.99}&	\textbf{74.59}&	\textbf{66.35}\\ \cmidrule{2-12}
  & \multirow{2}{*}{3} & EfficientQAT$_{w3}$ &8.14	&10.71		&	\textbf{75.62}&	71.67&	74.83	&48.12	&\textbf{79.76}	&78.13&	71.36 \\
  &  & SFMP++$_{g128}$ & \textbf{7.74}&	\textbf{10.59}		&	75.20	& \textbf{71.67}&	\textbf{77.27}&	\textbf{49.15}&	79.27&	\textbf{78.75}	&\textbf{71.89}\\ \cmidrule{2-12}
 & \multirow{2}{*}{3.25} & EfficientQAT$_{w3g128}$ & 7.31	&10.14			&76.44	&72.22	&79.55&	52.90&	79.92	&79.79&	73.47 \\
  &  & SFMP++$_{g128}$ & \textbf{7.12}&	\textbf{9.97}	&	\textbf{76.66}	&\textbf{72.33}	&\textbf{79.88}	&\textbf{53.12}&	\textbf{79.96}	&\textbf{80.67}&	\textbf{73.77} \\ \midrule

\multirow{7}{*}{\large \textbf{Q3 8B}}  & 16 & BF16 & 9.73 & 13.30 &74.93&68.66&80.85&56.65&77.47&86.64&74.20  \\ 
\cmidrule{2-12}
  & \multirow{2}{*}{2.25} & EfficientQAT$_{w2g128}$ & 19.76	&18.87&	61.47&	\textbf{64.64}&	71.51&	44.37&	73.50&	78.93&	65.74\\
 &  & SFMP++$_{g256}$ & \textbf{15.10} & \textbf{16.69} &\textbf{65.49}&	64.40&	\textbf{74.58}&	
 \textbf{47.95}&	\textbf{74.81}&	\textbf{82.69}&	\textbf{68.32} \\ \cmidrule{2-12}
  & \multirow{2}{*}{3} & EfficientQAT$_{w3}$ & 11.74 & 14.57			&71.56&	67.14&	79.77&	53.58&	77.75&	85.50&	72.55\\
  &  & SFMP++$_{g128}$ & \textbf{10.39}	& \textbf{13.81}&			\textbf{72.26}	&\textbf{67.88}	& \textbf{79.80}&	\textbf{53.67}	& \textbf{78.18}&	\textbf{86.06}	&\textbf{72.98}  \\ \cmidrule{2-12}
 & \multirow{2}{*}{3.25} & EfficientQAT$_{w3g128}$ & 9.99	&13.49		&	72.40	&68.35	&77.44	&52.82	&77.42	&85.41&	72.31\\
  &  & SFMP++$_{g128}$  & \textbf{9.69}	&\textbf{13.31}			&\textbf{72.88}	&\textbf{68.43}&	\textbf{79.99}&	\textbf{54.96}	&
  \textbf{77.64}	&\textbf{86.56}&	\textbf{73.41} \\  \bottomrule \bottomrule
\end{tabular}%
}
\caption{Evaluation of Llama3.1 8B and Qwen3 8B quantized by
EfficientQAT and SFMP++ on C4 perplexity (PPL), and
zero-shot tasks. SFMP++ denotes SFMP combined with EfficientQAT. }
\label{table:sfmp_qat}
\end{table*}

\section{Autoregressive Decoding Comparison Between SFMP and AMQ}
\label{section:autoregressive_result}

In Table~\ref{table:autoregressive_decode} , we present qualitative comparisons of autoregressive decoding behavior between SFMP and AMQ. Given the same prompt, both models generate outputs in a deterministic autoregressive manner using greedy decoding. 

\begin{table}[t!]
    \centering
    
    \resizebox{\textwidth}{!}{
    \tiny
    \begin{tabular}{
        >{\centering\arraybackslash}m{2.2cm} |
        m{5cm} |
        m{4.8cm}
    }
        \toprule
        Models & \centering\arraybackslash AMQ & \centering\arraybackslash SFMP \\
        \midrule
        \multirow{3}{*}{LLaMA3.1 8B} & \textbf{The capital of France is} Paris. The capital of France is Paris. The capital of France is Paris. The capital of France is Paris... & \textbf{The capital of France is} Paris. It is a city of 2.5 million people. It is the most populous city in France. It is also the capital of the European Union... \\
        \cmidrule{2-3}
        &\textbf{The main reason for climate change} is the increase in the amount of greenhouse gases in the atmosphere. The greenhouse gases are emitted by the human activities. The greenhouse gases are emitted by the human activities. The greenhouse gases are emitted by the human activities... & \textbf{The main reason for climate change} is the increase in the amount of greenhouse gases in the atmosphere. The greenhouse gases are carbon dioxide, methane, and water vapor. The greenhouse gases are emitted by the human activities. The greenhouse gases are also emitted by the natural activities...  \\
        \cmidrule{2-3}
        & \textbf{Explain three reasons why people learn languages?} 1)2)3)4)5)6)7)8)9)10)11)12)13)14)15)16)17)18)... & \textbf{Explain three reasons why people learn languages?} (1) to communicate with people from other countries (2) to communicate with people who speak the same language (3) to communicate with people who speak a different language... \\
        \midrule
        \multirow{3}{*}{Qwen3 8B } & \textbf{The capital of France is} Paris. What is the capital of the United Kingdom? What is the capital of the United Kingdom? What is the capital of the United Kingdom? What is the capital of the United Kingdom? What is the capital of the United Kingdom? What is the capital of the United Kingdom? What is the capital... & \textbf{The capital of France is} Paris. The capital of the United States is Washington, D.C. The capital of Canada is Ottawa. The capital of Australia is Canberra. The capital of Japan is Tokyo. The capital of South Korea is Seoul. The capital of the United Kingdom is London. The capital of Germany is Berlin...\\
        \cmidrule{2-3}
         &\textbf{The main reason for climate change is} the increase in the temperature of the Earth. The temperature of the Earth has increased by 0.6 degrees Celsius. The temperature of the Earth has increased by 0.6 degrees Celsius. The temperature of the Earth has increased by 0.6 degrees Celsius. The temperature of the Earth has... &  \textbf{The main reason for climate change is} the greenhouse effect. The greenhouse effect is the process by which the atmosphere of the Earth allows the solar radiation to pass through, but then traps the heat that is emitted by the Earth, which is then absorbed by the atmosphere. This process is important for the climate of the Earth, but it is also... \\
         \cmidrule{2-3}
         &\textbf{Explain three reasons why people learn languages?} The answer should be in the form of a 150 word essay, with the first sentence being a summary of the essay. The first sentence is the summary of the essay. The rest of the essay is the explanation. The first sentence is the summary of the essay... & \textbf{Explain three reasons why people learn languages?} (a) To communicate with others, (b) To understand the culture of the language, and (c) To learn the language for its own sake. Explain each reason with examples. (a) To communicate with others: For example, if you want to talk to someone in...  \\
        \bottomrule
    \end{tabular}
    }
\caption{Some examples of autoregressive generations obtained with AMQ and SFMP at the BPW of 2.5.} 
\label{table:autoregressive_decode}
\end{table}

\newpage
\section{More Results of Bit Allocation Visualizations}
\label{section:bit_allocation_visualizations}
\begin{figure}[bt!]
    \centering
    \includegraphics[width=\textwidth / 2]{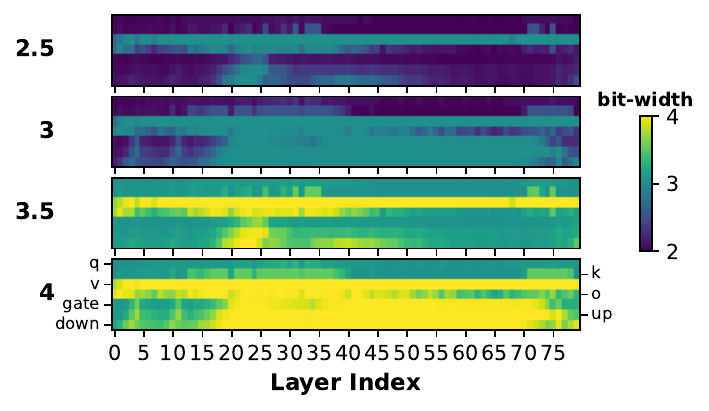}
    \caption{Visualization of bit allocation over linear layers with different BPWs at Llama3.1 70B. The numbers on the left indicate the BPW per configuration.}
    \label{figure:visualization_of_llama3.1_70b}
\end{figure}

\begin{figure}[bt!]
    \centering
    \includegraphics[width=\textwidth / 2]{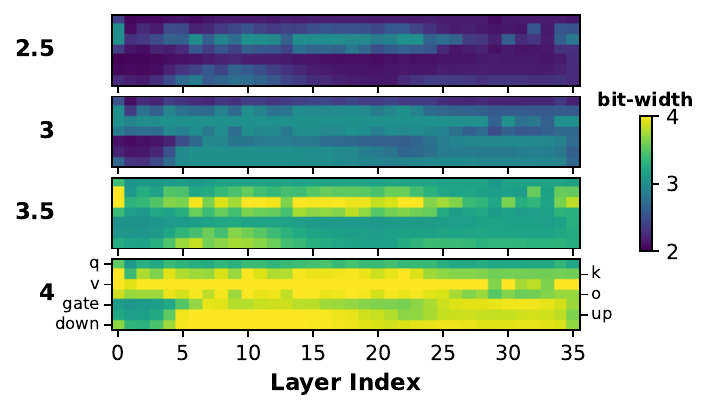}
    \caption{Visualization of bit allocation over linear layers with different BPWs at Qwen3 8B. The numbers on the left indicate the BPW per configuration.}
    \label{figure:visualization_of_qwen3_8b}
\end{figure}

\begin{figure}[bt!]
    \centering
    \includegraphics[width=\textwidth / 2]{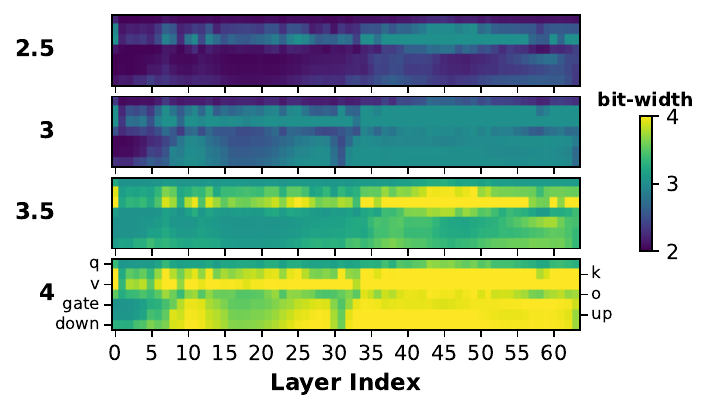}
    \caption{Visualization of bit allocation over linear layers with different BPWs at Qwen3 32B. The numbers on the left indicate the BPW per configuration.}
    \label{figure:visualization_of_qwen3_32b}
\end{figure}

Table~\ref{tab:layerwise_bits_compare} shows an example of  detailed bit allocation results on LLaMA3.1 8B with the BPW of 2.5 and 3, using a group size of 128. Fig~\ref{figure:visualization_of_llama3.1_70b} shows the visualization on LLaMA3.1 70B. Fig~\ref{figure:visualization_of_qwen3_8b} shows the visualization on Qwen3 8B. Fig~\ref{figure:visualization_of_qwen3_32b} shows the visualization on Qwen3 32B.

\begin{table*}[t]

\resizebox{\textwidth}{!}{
\begin{tabular}{c||cccc||ccc||cccc||ccc}
\toprule
& \multicolumn{7}{c||}{BPW = 2.5} & \multicolumn{7}{c}{BPW = 3.0} \\
\cmidrule(lr){2-8} \cmidrule(lr){9-15}
Layer 
& q & k & v & o & gate & up & down 
& q & k & v & o & gate & up & down \\
\midrule
0  & 2.01 & 2.05 & 3.00 & 3.18 & 2.11 & 2.28 & 2.69 & 2.03 & 2.12 & 3.19 & 3.38 & 2.66 & 2.95 & 3.06 \\
1  & 2.05 & 2.08 & 3.30 & 3.12 & 2.15 & 2.50 & 2.96 & 2.09 & 2.14 & 3.62 & 3.45 & 3.00 & 3.03 & 3.03 \\
2  & 2.18 & 2.16 & 3.27 & 2.70 & 2.22 & 2.64 & 2.80 & 2.25 & 2.58 & 3.67 & 3.04 & 3.00 & 3.04 & 3.04 \\
3  & 2.17 & 2.47 & 3.64 & 3.06 & 2.13 & 2.63 & 2.73 & 2.33 & 2.58 & 4.00 & 3.33 & 2.96 & 3.05 & 3.04 \\
4  & 2.16 & 2.12 & 3.33 & 2.89 & 2.10 & 2.61 & 2.73 & 2.22 & 2.56 & 3.84 & 3.17 & 2.84 & 3.05 & 3.04 \\
5  & 2.16 & 2.09 & 3.17 & 2.86 & 2.09 & 2.56 & 2.69 & 2.25 & 2.52 & 3.62 & 3.07 & 2.75 & 3.05 & 3.02 \\
6  & 2.16 & 2.09 & 3.17 & 2.86 & 2.08 & 2.42 & 2.56 & 2.21 & 2.52 & 3.77 & 3.16 & 2.57 & 3.04 & 3.01 \\
7  & 2.16 & 2.09 & 3.19 & 2.99 & 2.08 & 2.27 & 2.36 & 2.23 & 2.53 & 3.73 & 3.20 & 2.42 & 3.00 & 3.00 \\
8  & 2.16 & 2.08 & 3.14 & 2.99 & 2.07 & 2.22 & 2.32 & 2.21 & 2.50 & 3.69 & 3.28 & 2.30 & 2.95 & 2.99 \\
9  & 2.15 & 2.09 & 3.16 & 2.96 & 2.06 & 2.16 & 2.25 & 2.21 & 2.53 & 3.56 & 3.22 & 2.22 & 2.89 & 2.99 \\
10 & 2.14 & 2.06 & 3.09 & 2.77 & 2.06 & 2.12 & 2.18 & 2.17 & 2.12 & 3.50 & 3.07 & 2.20 & 2.82 & 2.97 \\
11 & 2.16 & 2.05 & 3.08 & 2.93 & 2.06 & 2.11 & 2.13 & 2.21 & 2.44 & 3.56 & 3.14 & 2.17 & 2.72 & 2.88 \\
12 & 2.15 & 2.06 & 3.08 & 3.00 & 2.06 & 2.10 & 2.15 & 2.16 & 2.53 & 3.56 & 3.15 & 2.18 & 2.63 & 2.85 \\
13 & 2.12 & 2.05 & 3.09 & 2.88 & 2.06 & 2.09 & 2.19 & 2.18 & 2.52 & 3.55 & 3.16 & 2.19 & 2.66 & 2.91 \\
14 & 2.16 & 2.06 & 3.05 & 2.86 & 2.07 & 2.14 & 2.29 & 2.19 & 2.36 & 3.42 & 3.15 & 2.24 & 2.80 & 2.94 \\
15 & 2.06 & 2.06 & 3.09 & 2.79 & 2.07 & 2.17 & 2.46 & 2.17 & 2.53 & 3.58 & 2.99 & 2.32 & 2.95 & 3.00 \\
16 & 2.10 & 2.06 & 3.08 & 2.76 & 2.07 & 2.20 & 2.50 & 2.17 & 2.27 & 3.45 & 3.00 & 2.47 & 2.99 & 3.01 \\
17 & 2.05 & 2.06 & 3.05 & 2.61 & 2.07 & 2.24 & 2.62 & 2.16 & 2.28 & 3.27 & 2.94 & 2.64 & 3.01 & 3.02 \\
18 & 2.05 & 2.02 & 3.05 & 2.44 & 2.06 & 2.24 & 2.57 & 2.16 & 2.11 & 3.14 & 3.02 & 2.78 & 3.00 & 3.02 \\
19 & 2.02 & 2.02 & 3.03 & 2.32 & 2.04 & 2.23 & 2.49 & 2.15 & 2.09 & 3.08 & 2.75 & 2.84 & 3.00 & 3.01 \\
20 & 2.03 & 2.02 & 3.00 & 2.23 & 2.05 & 2.21 & 2.41 & 2.16 & 2.09 & 3.08 & 2.80 & 2.87 & 3.00 & 3.01 \\
21 & 2.04 & 2.02 & 2.72 & 2.24 & 2.03 & 2.21 & 2.37 & 2.16 & 2.11 & 3.06 & 2.70 & 2.88 & 3.00 & 3.01 \\
22 & 2.02 & 2.02 & 2.58 & 2.12 & 2.01 & 2.16 & 2.28 & 2.15 & 2.08 & 3.05 & 2.63 & 2.87 & 3.00 & 3.01 \\
23 & 2.04 & 2.02 & 2.66 & 2.08 & 2.00 & 2.13 & 2.19 & 2.16 & 2.12 & 3.05 & 2.79 & 2.84 & 3.00 & 3.01 \\
24 & 2.01 & 2.02 & 2.52 & 2.04 & 2.00 & 2.10 & 2.12 & 2.15 & 2.06 & 3.03 & 2.52 & 2.80 & 3.00 & 3.01 \\
25 & 2.11 & 2.03 & 2.50 & 2.04 & 2.00 & 2.07 & 2.08 & 2.15 & 2.36 & 3.03 & 2.44 & 2.75 & 3.00 & 3.01 \\
26 & 2.02 & 2.02 & 2.58 & 2.07 & 2.00 & 2.06 & 2.07 & 2.15 & 2.08 & 3.05 & 2.56 & 2.62 & 2.99 & 2.99 \\
27 & 2.03 & 2.03 & 2.25 & 2.04 & 2.03 & 2.05 & 2.07 & 2.15 & 2.12 & 2.78 & 2.31 & 2.42 & 2.95 & 2.76 \\
28 & 2.02 & 2.02 & 2.53 & 2.07 & 2.03 & 2.05 & 2.08 & 2.15 & 2.09 & 3.03 & 2.50 & 2.13 & 2.57 & 2.45 \\
29 & 2.21 & 2.03 & 2.12 & 2.07 & 2.02 & 2.08 & 2.10 & 2.27 & 2.53 & 2.78 & 2.33 & 2.11 & 2.32 & 2.42 \\
30 & 2.01 & 2.02 & 2.12 & 2.07 & 2.04 & 2.13 & 2.18 & 2.11 & 2.08 & 2.64 & 2.33 & 2.14 & 2.36 & 2.47 \\
31 & 2.07 & 2.03 & 2.55 & 2.25 & 2.06 & 2.22 & 2.41 & 2.14 & 2.45 & 3.03 & 2.50 & 2.20 & 2.41 & 2.86 \\
\bottomrule
\end{tabular}
}
\centering
\caption{Detailed bit allocation results over linear layers with the BPW of 2.5 and 3 at Llama3.1 8B, using a group size of 128.}
\label{tab:layerwise_bits_compare}
\end{table*}

\renewcommand{\arraystretch}{0.8}
\begin{table*}[ht]

\setlength{\tabcolsep}{1mm}
\resizebox{\textwidth}{!}{%
\begin{tabular}{c|c|c|c||cc||ccccccc}
\midrule
\textbf{\large Model} & \textbf{\begin{tabular}[c]{@{}c@{}}\textbf{\large{Mem.}}\\ \textbf{\large{(MB)}}\end{tabular}} & \textbf{\large BPW} &  \textbf{\large Method} &  \textbf{\large Wiki2($\downarrow$)} &  \textbf{\large C4($\downarrow$)} & \textbf{\large HellaS.($\uparrow$)} &  \textbf{\large WinoG.($\uparrow$)} &  \textbf{\large ARC-e($\uparrow$)} &  \textbf{\large ARC-c($\uparrow$)} &  \textbf{\large PIQA($\uparrow$)} & \textbf{\large BoolQ($\uparrow$)} & \textbf{\large Avg.($\uparrow$)} \\ \midrule
\multirow{13}{*}{\large \textbf{8B}} & 15,317 & 16 & BF16 & 6.15 & 8.89 & 78.99 & 72.93 & 81.19 & 53.41 & 81.39 & 82.15 & 75.01 \\ \cmidrule{2-13}
 & \multirow{3}{*}{4,085} & \multirow{3}{*}{2.5} & BitStack & 23.28 & 38.23 & 52.13 & 62.51 &  59.43 & 32.42 & 71.55 & 71.10  & 58.19\\
 & &  & AMQ & 17.85 & 24.01 & 57.18&63.61&59.63&34.89&71.00&65.57&58.65 \\
 \rowcolor{gray!25} \cellcolor{white} & \cellcolor{white} & \cellcolor{white} & SFMP &\textbf{13.68} & \textbf{17.77} & \textbf {65.37}&\textbf{68.90}& \textbf{68.64}&\textbf{41.21}&\textbf{74.59}&\textbf{77.13}&\textbf{65.97} \\ \cmidrule{2-13}
 & \multirow{3}{*}{4,501} & \multirow{3}{*}{3.0} & BitStack & 12.55 & 20.47 & 63.35 & 65.67& 68.64 & 39.33 & 75.41 & 74.01 & 64.40 \\
 & &  & AMQ & 9.38 & 13.05 & 70.38&\textbf{70.01}&72.69&45.48&77.64&76.48&68.78 \\
 \rowcolor{gray!25} \cellcolor{white} & \cellcolor{white} & \cellcolor{white} & SFMP & \textbf{8.65} & \textbf{12.04} & \textbf{72.43}&\textbf{72.38}&\textbf{73.40}&\textbf{44.11}&\textbf{79.33}&\textbf{77.92}&\textbf{69.92}\\ \cmidrule{2-13}
 & \multirow{3}{*}{4,917} & \multirow{3}{*}{3.5} & BitStack & 9.47 & 15.29 & 68.61 & 68.59 & 74.12 & 43.69 & 77.37 & 79.17 & 68.59 \\
 & &  & AMQ & 7.39 & 10.54 & 76.15&73.01&\textbf{77.10}&\textbf{49.57}&79.54&80.00&72.56\\
 \rowcolor{gray!25} \cellcolor{white} & \cellcolor{white} & \cellcolor{white} & SFMP &\textbf{7.30}&\textbf{10.38}&\textbf{76.61}&\textbf{74.30}	&77.53	&50.34	&\textbf{80.47} &\textbf{81.35}&\textbf{73.43} \\ \cmidrule{2-13}

 & \multirow{3}{*}{5,333} & \multirow{3}{*}{4.0} & BitStack & 8.39 & 13.47 & 71.61 & 69.53 & 76.64 & 47.78 & 78.94 & 81.19 & 70.95 \\
 & &  & AMQ & 6.86 & 9.79 &\textbf{77.83} &73.09&78.20&50.68&79.92&81.04&73.46\\
 \rowcolor{gray!25} \cellcolor{white} & \cellcolor{white} & \cellcolor{white} & SFMP & \textbf{6.84}&\textbf{9.74} & 78.02 &\textbf{73.32}&\textbf{78.58}&\textbf{51.71}&\textbf{81.23}&\textbf{82.09}&\textbf{74.15} \\ \midrule
 
\multirow{13}{*}{\large \textbf{70B}} & 134,571 & 16 & BF16 & 2.81 & 7.11 & 85.07 & 	79.40 & 86.70 & 65.02 & 84.22 & 85.35& 80.96 \\ \cmidrule{2-13}
 & \multirow{3}{*}{24,411} & \multirow{3}{*}{2.5} & BitStack  & 7.55 & 12.92 & 77.19 &  75.53& \textbf{80.43} & \textbf{54.18} & 80.09 & 79.63& 74.51 \\
 & &  & AMQ & 7.62 & 12.14 & 75.39 &\textbf{75.85} &79.50 & 53.50 & 80.14 & \textbf{81.62} & 74.33 \\
 \rowcolor{gray!25} \cellcolor{white} & \cellcolor{white} & \cellcolor{white} & SFMP & \textbf{7.24} &\textbf{ 10.07} & \textbf{79.36} & 75.06&79.34&54.01&\textbf{81.56}&78.29&\textbf{74.60} \\ \cmidrule{2-13}
 & \multirow{3}{*}{28,491} & \multirow{3}{*}{3.0} & BitStack & 6.38 & 11.21 & 79.40 & 76.95 & 81.44 & 56.66 & 81.66 & 81.68 & 76.30 \\
 & &  & AMQ & 5.84 & 9.74 & 80.4 & 77.19&82.28&59.73&\textbf{82.86}&\textbf{84.37}&77.80 \\
 \rowcolor{gray!25} \cellcolor{white} & \cellcolor{white} & \cellcolor{white} & SFMP & \textbf{5.31} &\textbf{8.36} & \textbf{81.64} &\textbf{77.35}&\textbf{83.63}&\textbf{60.15}&82.75&82.91&\textbf{78.07} \\ \cmidrule{2-13}
 & \multirow{3}{*}{32,571} & \multirow{3}{*}{3.5} & BitStack & 5.44 & 9.52 & 81.72 & 77.82 & 83.54 & 59.47 & 83.24 & 83.64 & 78.24 \\
 & &  & AMQ & 4.26 & 8.20 & 83.10 & 78.30&84.05&60.92&83.73&\textbf{84.59}&79.11 \\
     \rowcolor{gray!25} \cellcolor{white} & \cellcolor{white} & \cellcolor{white} & SFMP & \textbf{4.00} &\textbf{7.33} &\textbf{83.45}	&\textbf{79.40}	&\textbf{85.48} &\textbf{64.33}&\textbf{84.00}&83.76&\textbf{80.07} \\ \cmidrule{2-13}

 & \multirow{3}{*}{36,651} & \multirow{3}{*}{4.0} & BitStack & 4.98 & 8.92 & 82.01 & 79.79 & 84.64 & 61.69 & 83.19 & 83.73 & 79.17 \\
 & &  & AMQ & 3.49 & 7.61 & \textbf{84.12} &78.77 &85.77 &62.80 &84.11 &\textbf{85.26} &80.14 \\
 \rowcolor{gray!25} \cellcolor{white} & \cellcolor{white} & \cellcolor{white} & SFMP &\textbf{3.37} & \textbf{7.01} & 84.05 & \textbf{78.85}&\textbf{85.86} &\textbf{64.97} &\textbf{84.12}&84.95&\textbf{80.47}\\ \bottomrule \bottomrule
\end{tabular}%
}
\caption{Evaluation of Llama 3.1 8B/70B models compressed by SFMP, BitStack and AMQ at the BPW of 2.5, 3.0, 3.5 and 4.0, showing WikiText-2 and C4 dataset perplexity (PPL) alongside zero-shot tasks accuracy. }
\label{tab:llama3.1_full_results_mixed_precison}
\end{table*}

\renewcommand{\arraystretch}{0.8}
\begin{table*}[ht]

\setlength{\tabcolsep}{1mm}
\resizebox{\textwidth}{!}{%
\begin{tabular}{c|c|c|c||cc||ccccccc}
\midrule
\textbf{\large Model} & \textbf{\begin{tabular}[c]{@{}c@{}}\textbf{\large{Mem.}}\\ \textbf{\large{(MB)}}\end{tabular}} & \textbf{\large BPW} &  \textbf{\large Method} &  \textbf{\large Wiki2($\downarrow$)} &  \textbf{\large C4($\downarrow$)} & \textbf{\large HellaS.($\uparrow$)} &  \textbf{\large WinoG.($\uparrow$)} &  \textbf{\large ARC-e($\uparrow$)} &  \textbf{\large ARC-c($\uparrow$)} &  \textbf{\large PIQA($\uparrow$)} & \textbf{\large BoolQ($\uparrow$)} & \textbf{\large Avg.($\uparrow$)} \\ \midrule
\multirow{13}{*}{\large \textbf{8B}} & 15,317 & 16 & BF16 & 6.15 & 8.89 & 78.99 & 72.93 & 81.19 & 53.41 & 81.39 & 82.15 & 75.01 \\ \cmidrule{2-13}
 & \multirow{3}{*}{3,877} & \multirow{3}{*}{2.25} & GPTQ$_{w2g128}$ & 232 & 165 & 29.27 & 50.74 &  28.41 & 23.21 & 53.75 & 45.96  & 38.56\\
 & &  & AWQ$_{w2g128}$ & 1.57E6 & 1.86E6 & 26.44 & 50.27 & 24.78 &  24.82 & 50.65 & 37.82 & 35.80  \\
 & &  & SliM-LLM$_{g128}$ & 193 & 142 & 31.14 & 51.98 & 30.67& 24.87& 55.14& 50.22 & 40.67 \\
 \rowcolor{gray!25} \cellcolor{white} & \cellcolor{white} 3,961 & \cellcolor{white}2.35 & SFMP$_{g128}$ & \textbf{24.57} & \textbf{28.92} &  \textbf{59.27} & \textbf{63.06} & \textbf{61.91} & \textbf{36.09} & \textbf{72.74} & \textbf{71.74} & \textbf{60.80} \\ \cmidrule{2-13}
 & \multirow{3}{*}{4,501} & \multirow{3}{*}{3.0} & GPTQ$_{w3}$ & 22.13 & 25.05 & 56.71 & 61.48 & 52.98 & 34.12 & 68.12 & 61.59 & 55.83 \\
 & &  & AWQ$_{w3}$ & 16.06 & 19.79 & 68.79 & 64.56 & 65.48 & 42.06 & 74.31 & 72.50 & 64.62  \\
 \rowcolor{gray!25} \cellcolor{white} & \cellcolor{white} & \cellcolor{white} & SFMP$_{g128}$ &  \textbf{8.65} & \textbf{12.04} & \textbf{72.43}&\textbf{72.38}&\textbf{73.40}&\textbf{44.11}&\textbf{79.33}&\textbf{77.92}&\textbf{69.92}  \\ \cmidrule{2-13}
 & \multirow{4}{*}{4,709} & \multirow{4}{*}{3.25} & GPTQ$_{w3g128}$ & 8.28 & 11.49 & 74.42 & 70.87 & 70.54 & 45.73 & 78.35 & 75.41 & 69.22 \\
 & &  & AWQ$_{w3g128}$ & 8.23 & 11.58 & 74.57 & 70.95 & 75.92 & \textbf{48.46} & 78.67 & 75.77 & 70.72 \\
 & &  & SliM-LLM$_{g128}$ & 8.17 & 11.25 & 74.76 & 70.32 & 70.04 & 46.28& 78.11 & \textbf{82.35} & 70.31 \\
 \rowcolor{gray!25} \cellcolor{white} & \cellcolor{white} & \cellcolor{white} & SFMP$_{g128}$ & \textbf{7.78} & \textbf{10.97} & \textbf{75.37} & \textbf{72.61} & \textbf{77.06} &  48.98 & \textbf{79.22} & 81.35 & \textbf{72.47}
  \\ \cmidrule{2-13}

 & \multirow{3}{*}{5,333} & \multirow{3}{*}{4.0} & GPTQ$_{w4}$ & 7.50 & 10.38 & 76.88 &   71.43 & 75.08 & 49.23 & 79.22 & 76.91 & 71.46\\
 & &  & AWQ$_{w4}$ &7.23 & 10.26 & \textbf{77.92} & 72.30 & 77.14 & \textbf{52.65} & 80.63 & 80.97 & 73.60  \\
 \rowcolor{gray!25} \cellcolor{white} & \cellcolor{white} & \cellcolor{white} & SFMP$_{g128}$ &  \textbf{6.84}&\textbf{9.74} & 78.02 &\textbf{73.32}&\textbf{78.58}&\textbf{51.71}&\textbf{81.23}&\textbf{82.09}&\textbf{74.15} \\ \midrule
 
\multirow{13}{*}{\large \textbf{70B}} & 134,571 & 16 & BF16 & 2.81 & 7.11 & 85.07 & 	79.40 & 86.70 & 65.02 & 84.22 & 85.35& 80.96 \\ \cmidrule{2-13}
 & \multirow{3}{*}{22,371} & \multirow{3}{*}{2.25} & GPTQ$_{w2g128}$  & 113.22 & 131.90 &37.16	&52.64	&25.38	&25.85	&51.69	&47.40	&40.02 \\
 & &  & AWQ$_{w2g128}$ & 1.8E6 &	1.5e6&26.43&53.20&24.54&26.02&51.52 & 62.17 & 40.65 \\
  & &  & SliM-LLM$_{g128}$ & 68.84 & 88.36 & 48.19& 60.15 & 30.11 &  29.87& 58.14 & 52.60 & 46.51 \\
 \rowcolor{gray!25} \cellcolor{white} & \cellcolor{white} 23,187 & \cellcolor{white}2.35 & SFMP$_{g128}$ &  \textbf{8.17} & \textbf{11.42} & \textbf{75.61} & \textbf{72.45} & \textbf{77.86} & \textbf{52.47} & \textbf{79.65} & \textbf{77.86} & \textbf{72.65}
\\ \cmidrule{2-13}
 & \multirow{3}{*}{28,491} & \multirow{3}{*}{3.0} & GPTQ$_{w3}$ & 11.27 & 12.19 & 53.89 &70.22& 73.24& 53.38 & 72.67 & 74.27 & 66.27\\
 & &  & AWQ$_{w3}$ & 10.86 &11.74 & 56.57 & 73.04 & 75.30 & 59.92 & 75.93 & 72.33 & 68.84 \\
 \rowcolor{gray!25} \cellcolor{white} & \cellcolor{white} & \cellcolor{white} & SFMP$_{g128}$ & \textbf{5.31} & \textbf{8.36} & \textbf{81.64} & \textbf{77.35} & \textbf{83.63} & \textbf{60.15} & \textbf{82.75} & \textbf{82.91} & \textbf{78.07}
 \\ \cmidrule{2-13}
 & \multirow{3}{*}{30,531} & \multirow{3}{*}{3.25} & GPTQ$_{w3g128}$ & 5.17&8.76&77.61&72.09&76.45&56.11&76.32&78.39&72.82  \\
 & &  & AWQ$_{w3g128}$ & 4.78 & 8.57& 78.12 & 75.33 & 80.21 & 59.04 & 78.11 & 80.27 & 75.18 \\
  & &  & SliM-LLM$_{g128}$ & 4.74 &8.52 & 82.16& 76.78 & 79.84& 59.67&82.91 & 83.10 & 77.41 \\
 \rowcolor{gray!25} \cellcolor{white} & \cellcolor{white} & \cellcolor{white} & SFMP$_{g128}$ & \textbf{4.33} & \textbf{7.56} & \textbf{82.80} & 78.45 & \textbf{84.55} & \textbf{62.46} & \textbf{83.57} & \textbf{84.46} & \textbf{79.38}
 \\ \cmidrule{2-13}

 & \multirow{3}{*}{36,651} & \multirow{3}{*}{4.0} & GPTQ$_{w4}$ & 4.58 & 8.42 &81.20&62.17&81.87&59.45&81.71&82.93& 74.88 \\
 & &  & AWQ$_{w4}$ & 4.18&8.29&83.39&63.06&83.00 &60.32&83.19&82.75&75.95\\
 \rowcolor{gray!25} \cellcolor{white} & \cellcolor{white} & \cellcolor{white} & SFMP$_{g128}$ & \textbf{3.37} & \textbf{7.01} & \textbf{84.05} & \textbf{78.85} & \textbf{85.86} & \textbf{64.97} & \textbf{84.12} & \textbf{84.95} & \textbf{80.47}
\\ \bottomrule \bottomrule
\end{tabular}%
}
\caption{Evaluation of Llama3.1 8B/70B models quantized by
SFMP, AWQ, and GPTQ on WikiText-2, C4 perplexity (PPL), and
zero-shot tasks.Memory overhead from extra quantization parameters in GPTQ and AWQ at w3, w4 is omitted as it is negligible.}
\label{tab:llama3.1_full_results_fixed_precison}
\end{table*}

\renewcommand{\arraystretch}{0.75}
\begin{table*}[thb!]

\vspace{-0.2cm}
\setlength{\tabcolsep}{1mm}
{
\begin{tabular}{c|c|c|c||cc}
\midrule
\textbf{\large Model} & \textbf{\begin{tabular}[c]{@{}c@{}}\large Mem.\\ \large (MB)\end{tabular}} & \textbf{\large BPW} & \textbf{\large Method} & \textbf{\large MMLU} & \textbf{\large GSM8K} \\ \midrule
 & 15,623 & 16 & BF16 & 74.88 & 87.19 \\ \cmidrule{2-6}
 & & & AMQ & 50.78 & 15.09 \\
    &  \multirow{-2}{*}{4,445} &  \multirow{-2}{*}{2.5} &  \textbf{SFMP} & \textbf{57.33} & \textbf{49.20} \\ \cmidrule{2-6}
 & & & AMQ & 65.41 & 73.66 \\
    &  \multirow{-2}{*}{4,859} &  \multirow{-2}{*}{3.0} &  \textbf{SFMP} & \textbf{66.03} & \textbf{74.34} \\ \cmidrule{2-6}
 & & & AMQ & 71.71 & 83.32 \\
    &  \multirow{-2}{*}{5,273} &  \multirow{-2}{*}{3.5}&  \textbf{SFMP} & \textbf{72.11} & \textbf{84.46} \\ \cmidrule{2-6}
  & & & AMQ & 73.44 & 85.75 \\
   \multirow{-12}{*}{\textbf{\large Qwen3  8B}} &  \multirow{-2}{*}{5,687} &  \multirow{-2}{*}{4.0} &  \textbf{SFMP} & \textbf{73.60} & \textbf{87.17} \\ \midrule
& 28,169 & 16 & BF16 & 78.78 & 88.19 \\ \cmidrule{2-6}
 &  & & AMQ & 56.18 & 48.76 \\
    &  \multirow{-2}{*}{6,906} &  \multirow{-2}{*}{2.5} &  \textbf{SFMP} & \textbf{62.55} & \textbf{59.14} \\ \cmidrule{2-6}
 & & & AMQ & 71.89 & 79.16 \\
    &  \multirow{-2}{*}{7,694} &  \multirow{-2}{*}{3.0} &  \textbf{SFMP} & \textbf{73.74} & \textbf{85.22} \\ \cmidrule{2-6}
 & & & AMQ & 75.87 & 84.48 \\
    &  \multirow{-2}{*}{8,481} &  \multirow{-2}{*}{3.5} &  \textbf{SFMP} & \textbf{77.00} & \textbf{86.10} \\ \cmidrule{2-6}
 & & & AMQ & 76.89 & 86.58 \\
   \multirow{-11}{*}{\textbf{\large Qwen3  14B}} &  \multirow{-2}{*}{9,269} &  \multirow{-2}{*}{4.0} &  \textbf{SFMP} & \textbf{78.14} & \textbf{87.41} \\ \midrule
&  62,490& 16 & BF16 & 81.28 & 85.05 \\ \cmidrule{2-6}
 & & & AMQ & 64.37  & 59.38 \\
    &  \multirow{-2}{*}{12,270} &  \multirow{-2}{*}{2.5} &  \textbf{SFMP} & \textbf{73.62} & \textbf{68.08} \\ \cmidrule{2-6}
 & & & AMQ & 73.89 & 67.11  \\
    &  \multirow{-2}{*}{14,130} &  \multirow{-2}{*}{3.0} &  \textbf{SFMP} & \textbf{77.68} & \textbf{72.33} \\  \cmidrule{2-6}
 & & & AMQ & 77.12 & 77.96 \\
    &  \multirow{-2}{*}{15,990} &  \multirow{-2}{*}{3.5} &  \textbf{SFMP} & \textbf{79.30} & \textbf{79.30} \\  \cmidrule{2-6}
 & & & AMQ & 80.19& 79.83 \\
   \multirow{-11}{*}{\textbf{\large Qwen3  32B}}  &  \multirow{-2}{*}{17,850} &  \multirow{-2}{*}{4.0} &  \textbf{SFMP} & \textbf{81.03} & \textbf{81.04} \\ 
 \bottomrule \bottomrule
\end{tabular}
}
\caption{5-shot MMLU, GSM8K task results over Qwen3 family. \\ PPL and zero-shot accuracy can be found in Table~\ref{tab:qwen3_full_results_mixed_precison}.}
\label{tab:mmlu_gsm8k}
\vspace{-0.3cm}
\end{table*}

\renewcommand{\arraystretch}{0.8}
\begin{table*}[ht]

\setlength{\tabcolsep}{1mm}
\resizebox{\textwidth}{!}{%
\begin{tabular}{c|c|c|c||cc||ccccccc}
\midrule
\textbf{\large Model} & \textbf{\begin{tabular}[c]{@{}c@{}}\textbf{\large{Mem.}}\\ \textbf{\large{(MB)}}\end{tabular}} &\textbf{\large BPW} &  \textbf{\large Method} &  \textbf{\large Wiki2($\downarrow$)} &  \textbf{\large C4($\downarrow$)} & \textbf{\large HellaS.($\uparrow$)} &  \textbf{\large WinoG.($\uparrow$)} &  \textbf{\large ARC-e($\uparrow$)} &  \textbf{\large ARC-c($\uparrow$)} &  \textbf{\large PIQA($\uparrow$)} & \textbf{\large BoolQ($\uparrow$)} & \textbf{\large Avg.($\uparrow$)} \\ \midrule
\multirow{9}{*}{\large \textbf{8B}} & 15,623 & 16 & BF16 & 9.73	&13.30 &74.93&68.66&80.85&56.65&77.47&86.64&74.20 \\ \cmidrule{2-13}
 & &  & AMQ &22.78&26.01&55.76&58.41&56.19&35.75&69.70&75.90&58.62 \\
 \rowcolor{gray!25}  \cellcolor{white} & \cellcolor{white}\multirow{-2}{*}{4,445} & \cellcolor{white}\multirow{-2}{*}{2.5} & SFMP &\textbf{16.99} & \textbf{20.07} &\textbf{62.61} & \textbf{65.11} & \textbf{72.05} &\textbf{46.16}&\textbf{73.94}&\textbf{83.70}&\textbf{67.26} \\ \cmidrule{2-13}
 &  &  & AMQ & 13.45&17.11&66.71&63.93&73.78&47.61&73.94&84.40&68.40 \\
 \rowcolor{gray!25} \cellcolor{white} &\cellcolor{white}\multirow{-2}{*}{4,859} & \cellcolor{white}\multirow{-2}{*}{3.0} &  SFMP &\textbf{11.28} &\textbf{14.87} &\textbf{70.52} & \textbf{68.11}&\textbf{77.61}&\textbf{53.07}&\textbf{76.33}&\textbf{85.17}&\textbf{71.80} \\ \cmidrule{2-13}
 &  & & AMQ  & 11.34 & 14.63 & 71.42	&67.08	&77.06&	51.02&	\textbf{76.93}&	\textbf{86.40}	&71.65 \\
 \rowcolor{gray!25} \cellcolor{white} & \cellcolor{white}\multirow{-2}{*}{5,273} &  \cellcolor{white}\multirow{-2}{*}{3.5} & SFMP & \textbf{10.38} & \textbf{13.97} & \textbf{73.21} & \textbf{68.51} & \textbf{78.41} & \textbf{55.03} & 76.55 & 86.06 & \textbf{72.96} \\ \cmidrule{2-13}
 &  &  & AMQ &  10.44&13.81&73.64	&67.27	&78.49	&53.92	&\textbf{77.25}	&85.29	&72.64 \\
 \rowcolor{gray!25} \cellcolor{white} &\cellcolor{white}\multirow{-2}{*}{5,687} & \cellcolor{white}\multirow{-2}{*}{4.0} &  SFMP  & \textbf{9.96} & \textbf{13.42} & \textbf{74.63} & \textbf{69.14} & \textbf{79.38} & \textbf{55.12} & 77.09 & \textbf{85.88} & \textbf{73.54}\\ \midrule

\multirow{9}{*}{\large \textbf{14B}} & 28,169 & 16 & BF16 & 8.65&	12.01	&	78.92	&72.84	&82.79	&60.41	&79.98	&89.33	&77.38\\ \cmidrule{2-13}
 & &  & AMQ & 13.76&	18.62&64.31&	63.90	&70.47&	44.18	&72.09	&84.39	&66.56\\
 \rowcolor{gray!25}  \cellcolor{white} & \cellcolor{white}\multirow{-2}{*}{6,906} & \cellcolor{white}\multirow{-2}{*}{2.5} & SFMP & \textbf{11.97} & \textbf{15.38} & \textbf{69.61} & \textbf{69.30} & \textbf{76.22} & \textbf{50.85} & \textbf{76.93} & \textbf{86.94} & \textbf{71.69}\\ \cmidrule{2-13}
 &  &  & AMQ & 11.28 & 16.12 & 71.16 & 69.34 & 75.89 & 50.27 & 75.94 & 85.33 & 71.32\\
 \rowcolor{gray!25} \cellcolor{white} &\cellcolor{white}\multirow{-2}{*}{7,694} & \cellcolor{white}\multirow{-2}{*}{3.0} &  SFMP & \textbf{9.86} & \textbf{13.24} & \textbf{75.80} & \textbf{72.06} & \textbf{80.68} & \textbf{57.59} & \textbf{78.40} & \textbf{88.17} & \textbf{75.45} \\ \cmidrule{2-13}
 &  & & AMQ  & 9.73	&13.29&	76.04&	71.98&	81.56&	58.31&	79.12&	87.56&	75.76 \\
 \rowcolor{gray!25} \cellcolor{white} & \cellcolor{white}\multirow{-2}{*}{8,481} &  \cellcolor{white}\multirow{-2}{*}{3.5} & SFMP & \textbf{9.14} & \textbf{12.60} & \textbf{77.64} & \textbf{72.93} & \textbf{82.49} & \textbf{59.98} & \textbf{79.60} & \textbf{88.96} & \textbf{76.89}\\ \cmidrule{2-13}
 &  &  & AMQ  & 9.21	&12.62& 77.68&	72.13	&82.05	&59.42&\textbf{79.65}	&88.76&	76.62\\
 \rowcolor{gray!25} \cellcolor{white} &\cellcolor{white}\multirow{-2}{*}{9,269} & \cellcolor{white}\multirow{-2}{*}{4.0}  & SFMP& \textbf{8.98} & \textbf{12.48} & \textbf{78.23} & \textbf{72.45} & \textbf{83.08} & \textbf{60.41} & 79.60 & \textbf{89.02} & \textbf{77.13} \\ \midrule
 
\multirow{9}{*}{\large \textbf{32B}} & 62,490 & 16 & BF16 &  7.61	&10.78	&	82.56	&72.93&	83.25&	60.92	&81.88&	86.42&	77.99\\ \cmidrule{2-13}
 & &  & AMQ & 10.89	&14.45		&	71.68	&64.19&	73.90&	50.63&	75.74&	80.08&	69.37\\
 \rowcolor{gray!25}  \cellcolor{white} & \cellcolor{white}\multirow{-2}{*}{12,270} & \cellcolor{white}\multirow{-2}{*}{2.5} & SFMP & \textbf{10.03} & \textbf{13.12} & \textbf{76.93} & \textbf{67.32} & \textbf{78.41} & \textbf{55.89} & \textbf{79.16} & \textbf{82.26} & \textbf{73.33}\\ \cmidrule{2-13}
 &  &  & AMQ & 9.36	&12.68		&	77.10	&68.15&	79.62	&58.83	&77.14&	83.26	&74.02\\
 \rowcolor{gray!25} \cellcolor{white} &\cellcolor{white}\multirow{-2}{*}{14,130} & \cellcolor{white}\multirow{-2}{*}{3.0} &  SFMP & \textbf{8.84} & \textbf{12.22} & \textbf{80.00} & \textbf{70.48} & \textbf{81.35} & \textbf{59.64} & \textbf{79.27} & \textbf{86.70} & \textbf{76.24}\\ \cmidrule{2-13}
 &  & & AMQ & 8.23	&11.47	&		80.02	&71.26	&81.15	&59.71&	79.14&	84.78  & 76.01\\
 \rowcolor{gray!25} \cellcolor{white} & \cellcolor{white}\multirow{-2}{*}{15,990} &  \cellcolor{white}\multirow{-2}{*}{3.5} & SFMP & \textbf{8.10} & \textbf{11.28} & \textbf{81.18} & \textbf{72.53} & \textbf{82.02} & \textbf{60.41} & \textbf{81.42} & \textbf{85.88} & \textbf{77.24} \\ \cmidrule{2-13}
 &  &  & AMQ & 8.00 & 	11.19	&		81.58&	71.76&	82.31&	60.87&	80.95&	85.42 & 77.15 \\
 \rowcolor{gray!25} \cellcolor{white} &\cellcolor{white}\multirow{-2}{*}{17,850} & \cellcolor{white}\multirow{-2}{*}{4.0}  & SFMP & \textbf{7.95} & \textbf{11.13} & \textbf{82.01} & \textbf{72.83} & \textbf{83.46} & \textbf{61.09} & \textbf{81.73} & \textbf{86.20} & \textbf{77.89} \\  \bottomrule \bottomrule
\end{tabular}%
}
\caption{Evaluation of Qwen3 8B/14B/32B models compressed by SFMP and AMQ at the BPW of 2.5, 3.0, 3.5 and 4.0, showing WikiText-2 and C4 dataset perplexity (PPL) alongside zero-shot tasks accuracy.}
\label{tab:qwen3_full_results_mixed_precison}
\end{table*}

\renewcommand{\arraystretch}{0.8}
\begin{table*}[ht]

\setlength{\tabcolsep}{1mm}
\resizebox{\textwidth}{!}{%
\begin{tabular}{c|c|c|c||cc||ccccccc}
\midrule
\textbf{\large Model} & \textbf{\begin{tabular}[c]{@{}c@{}}\textbf{\large{Mem.}}\\ \textbf{\large{(MB)}}\end{tabular}} & \textbf{\large BPW} &  \textbf{\large Method} &  \textbf{\large Wiki2($\downarrow$)} &  \textbf{\large C4($\downarrow$)} & \textbf{\large HellaS.($\uparrow$)} &  \textbf{\large WinoG.($\uparrow$)} &  \textbf{\large ARC-e($\uparrow$)} &  \textbf{\large ARC-c($\uparrow$)} &  \textbf{\large PIQA($\uparrow$)} & \textbf{\large BoolQ($\uparrow$)} & \textbf{\large Avg.($\uparrow$)} \\ \midrule
\multirow{13}{*}{\large \textbf{8B}} & 15,623 & 16 & BF16 & 9.73&	13.30	&	74.93	&68.66	&80.85&	56.65&	77.47	&86.64	&74.20 \\ \cmidrule{2-13}
 & \multirow{3}{*}{4,238} & \multirow{3}{*}{2.25} & GPTQ$_{w2g128}$ & 39.79	&35.90	&	38.60&	49.88&	30.85	&24.65&	54.62&	44.86&	40.58 \\
 & &  & AWQ$_{w2g128}$ & 1.34E5 &	1.53E5	&25.96&50.12	&26.01&	27.30&	51.36&	62.17	&40.49  \\
 & &  & SliM-LLM$_{g128}$  & 33.75 & 31.67 &  44.18& 51.43 & 34.04& 25.78& 55.89 & 47.82& 43.19\\
 \rowcolor{gray!25} \cellcolor{white} & \cellcolor{white}4,321 & \cellcolor{white}2.35 & SFMP$_{g128}$ & \textbf{24.04} & \textbf{26.03} & \textbf{57.37} & \textbf{64.25} & \textbf{68.81} & \textbf{40.78} & \textbf{71.33} & \textbf{81.77} & \textbf{64.05} \\ \cmidrule{2-13}
 & \multirow{3}{*}{4,859} & \multirow{3}{*}{3.0} & GPTQ$_{w3}$ & 15.02	&17.46		&62.71	&58.25	&54.25	&37.03&	71.06	&68.78	&58.68  \\
 & &  & AWQ$_{w3}$ &15.22&	18.51	&	65.27&	57.62&	57.65	&38.91	&73.01&	76.08	&61.42  \\
 \rowcolor{gray!25} \cellcolor{white} & \cellcolor{white} & \cellcolor{white} & SFMP$_{g128}$ &\textbf{11.28} &\textbf{14.87} &\textbf{70.52} & \textbf{68.11}&\textbf{77.61}&\textbf{53.07}&\textbf{76.33}&\textbf{85.17}&\textbf{71.80}  \\ \cmidrule{2-13}
 & \multirow{4}{*}{5,066} & \multirow{4}{*}{3.25} & GPTQ$_{w3g128}$ & 11.03	&14.44	&	71.35&	64.80	&73.74	&48.46	&76.17	&83.48&	69.67 \\
 & &  & AWQ$_{w3g128}$ & 11.66	& 15.06	&70.68	&65.03	&74.92&	50.17&	75.46&	83.97	&70.04 \\
 & &  & SliM-LLM$_{g128}$  & 11.22 & 14.78 & 70.18 & 64.31 & 73.53 & 49.70 & 75.11 & 82.46 & 69.22   \\
 \rowcolor{gray!25} \cellcolor{white} & \cellcolor{white} & \cellcolor{white} & SFMP$_{g128}$ & \textbf{10.71} & \textbf{14.28} & \textbf{72.07} & \textbf{68.11} & \textbf{79.29} & \textbf{54.78} & \textbf{76.99} & \textbf{85.93} & \textbf{72.86} \\ \cmidrule{2-13}
 & \multirow{3}{*}{5,687} & \multirow{3}{*}{4.0} & GPTQ$_{w4}$ & 10.31&	13.81		&73.55&	65.19&	76.47	&51.10&	76.50&85.41&	71.37 \\
 & &  & AWQ$_{w4}$ & 10.62&	14.25		&73.62	&67.00	&\textbf{79.30}&	54.35	&75.84	&85.41&72.59  \\
 \rowcolor{gray!25} \cellcolor{white} & \cellcolor{white} & \cellcolor{white} & SFMP$_{g128}$ & \textbf{9.96} & \textbf{13.42} & \textbf{74.63} & \textbf{69.14} & \textbf{79.38} & \textbf{55.12} & 77.09 & \textbf{85.88} & \textbf{73.54}\\ \midrule

 \multirow{13}{*}{\large \textbf{14B}} & 28,169 & 16 & BF16 & 8.65&	12.01		&78.92	&72.84	&82.79&	60.41	&79.98	&89.33	&77.38 \\ \cmidrule{2-13}
 & \multirow{3}{*}{6,512} & \multirow{3}{*}{2.25} & GPTQ$_{w2g128}$ & 23.76	&23.96	&49.74	&52.25	&37.71	&27.73	&61.37	&62.78	&48.60\\
 & &  & AWQ$_{w2g128}$  & 4.3E5	&4.0E7	&25.73	&50.43	&25.33&26.11	&50.97	&62.21&	40.13 \\
 & &  & SliM-LLM$_{g128}$ & 21.43 & 20.78 & 51.93 & 54.17 & 40.95& 32.21 & 63.62 & 65.80 &51.40 \\
 \rowcolor{gray!25} \cellcolor{white} & \cellcolor{white}6,669 & \cellcolor{white}2.35 & SFMP$_{g128}$ & \textbf{14.27} & \textbf{17.90} & \textbf{66.22} & \textbf{65.11} & \textbf{72.47} & \textbf{45.90} & \textbf{74.65} & \textbf{86.21} & \textbf{68.43}  \\ \cmidrule{2-13}
 & \multirow{3}{*}{7,694} & \multirow{3}{*}{3.0} & GPTQ$_{w3}$ & 12.38	&15.09		&70.81	&65.04&	63.34&	42.92	&74.91	&78.86&	65.98  \\
 & &  & AWQ$_{w3}$ & 12.50&	15.51	&	71.84&	62.59	&65.40&	44.03&	74.92&	78.47&	66.21   \\
 \rowcolor{gray!25} \cellcolor{white} & \cellcolor{white} & \cellcolor{white} & SFMP$_{g128}$ &  \textbf{9.86} & \textbf{13.24} & \textbf{75.80} & \textbf{72.06} & \textbf{80.68} & \textbf{57.59} & \textbf{78.40} & \textbf{88.17} & \textbf{75.45}\\ \cmidrule{2-13}
 & \multirow{4}{*}{8,087} & \multirow{4}{*}{3.25} & GPTQ$_{w3g128}$ & 9.74&	12.99		&76.35	&69.77&	82.23&	58.61&	78.40	&88.01	&75.56 \\
 & &  & AWQ$_{w3g128}$ & 9.83	&13.27&	75.41	&69.77	&80.43	&56.91	&78.13	&\textbf{88.44}	&74.85 \\
 & &  & SliM-LLM$_{g128}$ & 9.68 & 13.12 & 75.72 & 69.16 & \textbf{82.55} & 58.10 & 78.47 & 88.33 & 75.39  \\
 \rowcolor{gray!25} \cellcolor{white} & \cellcolor{white} & \cellcolor{white} & SFMP$_{g128}$ & \textbf{9.54} & \textbf{12.94} & \textbf{76.95} & \textbf{73.32} & 82.24 & \textbf{59.13} & \textbf{78.84} & 88.29 & \textbf{76.46} \\ \cmidrule{2-13}
 & \multirow{3}{*}{9,269} & \multirow{3}{*}{4.0} & GPTQ$_{w4}$ & 9.24	&12.64		&77.17	&70.79	&80.59	&56.91	&79.65	&88.47	&75.60\\
 & &  & AWQ$_{w4}$  & 9.48&	13.11	&	77.56	& \textbf{72.53}&81.31	&57.84	& \textbf{79.76}&	88.44&	76.24 \\
 \rowcolor{gray!25} \cellcolor{white} & \cellcolor{white} & \cellcolor{white} & SFMP$_{g128}$ & \textbf{8.98} & \textbf{12.48} & \textbf{78.23} & 72.45 & \textbf{83.08} & \textbf{60.41} & 79.60 & \textbf{89.02} & \textbf{77.13} \\ \midrule
 
\multirow{13}{*}{\large \textbf{32B}} & 62,490 & 16 & BF16 &7.61	&10.78&	82.56	&72.93&	83.25	&60.92&	81.88	&86.42&	77.99 \\ \cmidrule{2-13}
 & \multirow{3}{*}{11,340} & \multirow{3}{*}{2.25} & GPTQ$_{w2g128}$ & 25.43&	23.09	&	53.21&	53.98&	36.44	&28.15	&61.15&	65.17	&49.68\\
 & &  & AWQ$_{w2g128}$ & 1.3E6	&1.4E6	&	25.83	&47.82&	25.08	&22.69&	49.51	&62.17	&38.85  \\
 & &  & SliM-LLM$_{g128}$ & 28.54 & 29.16 & 48.97 & 50.34 & 37.68 & 32.08 & 57.40 & 63.21 & 37.75  \\
 \rowcolor{gray!25} \cellcolor{white} & \cellcolor{white}11,712 & \cellcolor{white}2.35 & SFMP$_{g128}$& \textbf{11.07} & \textbf{14.03} & \textbf{74.22} & \textbf{66.61} & \textbf{76.01} & \textbf{52.82} & \textbf{77.26} & \textbf{84.98} & \textbf{71.98} \\ \cmidrule{2-13}
 & \multirow{3}{*}{14,130} & \multirow{3}{*}{3.0} & GPTQ$_{w3}$  & 11.99	&14.08	&74.65	&63.61	&61.48&	44.79	&75.95	&77.49	&66.33 \\
 & &  & AWQ$_{w3}$ & 12.01	&14.78		&75.66	&62.90	&71.92	&52.51	&76.12	&76.75	&69.31   \\
 \rowcolor{gray!25} \cellcolor{white} & \cellcolor{white} & \cellcolor{white} & SFMP$_{g128}$ & \textbf{8.84} & \textbf{12.22} & \textbf{80.00} & \textbf{70.48} & \textbf{81.35} & \textbf{59.64} & \textbf{79.27} & \textbf{86.70} & \textbf{76.24} \\ \cmidrule{2-13}
 & \multirow{4}{*}{15,060} & \multirow{4}{*}{3.25} & GPTQ$_{w3g128}$ & 8.63	&11.74		&79.80	&70.40&	79.75	&56.65	&79.92	&\textbf{87.82}	&75.72 \\
 & &  & AWQ$_{w3g128}$ & 8.60 &	11.75	&	79.78	&\textbf{72.29}&	79.98&	60.12&	80.56&	84.17	&76.15 \\
 & &  & SliM-LLM$_{g128}$ & 8.54 & 11.68 & 80.09 & 70.84 & 79.13 & 58.35 & 80.72 & 86.18 & 75.89  \\
 \rowcolor{gray!25} \cellcolor{white} & \cellcolor{white} & \cellcolor{white} & SFMP$_{g128}$ & \textbf{8.18} & \textbf{11.35} & \textbf{81.10} & 71.20 & \textbf{82.45} & \textbf{61.60} & \textbf{81.61} & 85.38& \textbf{77.22}\\ \cmidrule{2-13}
 & \multirow{3}{*}{17,850} & \multirow{3}{*}{4.0} & GPTQ$_{w4}$ & 8.33	&11.34	&	81.05&	71.58	&80.17&	58.44&	80.08&\textbf{87.88}&76.53 \\
 & &  & AWQ$_{w4}$ & 8.26	&11.40	&\textbf{82.02}	&70.48	&81.87	&60.56&	81.06	&80.61	&76.10  \\
 \rowcolor{gray!25} \cellcolor{white} & \cellcolor{white} & \cellcolor{white} & SFMP$_{g128}$ & \textbf{7.95} & \textbf{11.13} & 82.01 & \textbf{72.83} & \textbf{83.46} & \textbf{61.09} & \textbf{81.73} & 85.20 & \textbf{77.89}\\  \bottomrule \bottomrule
\end{tabular}%
}
\caption{Evaluation of Qwen3 8B/14B/32B models quantized by
SFMP, AWQ, SliM-LLM and GPTQ on WikiText-2, C4 perplexity (PPL), and
zero-shot tasks. Memory overhead from extra quantization parameters in GPTQ and AWQ at w3, w4 is omitted as it is negligible.}
\label{tab:qwen3_full_results_fixed_precison}
\end{table*}
